%% file: TRC_KinectsAndHumanKinetics.tex
\documentclass[preprint,authoryear,12pt]{elsarticle}

\usepackage{graphicx}

\usepackage{amssymb}
\usepackage{amsthm}
\usepackage{amsmath}

\newcommand{\kinect}{{S_k}}
\newcommand{\trajectory}{{\cal T}}

\newcommand{\trajectoryGT}{{\cal T}_{{\scriptsize \mbox{G}}}}
\newcommand{\translation}{{\mathbf t}}
\newcommand{\rotation}{{\mathbf R}}

\newcommand{\depthpointset}{{\cal D}}
\newcommand{\depthpointsetprime}{{\cal D'}}
\newcommand{\depthpointsetdprime}{{\cal D''}}

\newcommand{\clusteridprime}{\depthpointset_i''}
\newcommand{\clusterjdprime}{\depthpointset_j''}

\newcommand{\ped}{{\mathbf x}_{{\scriptsize \mbox p}_{i}}}

\newcommand{\pedone}{{\mathbf x}_{{\scriptsize \mbox p}_{1}}}
\newcommand{\pedonex}{x_{{\scriptsize \mbox p}_1} }
\newcommand{\pedoney}{y_{{\scriptsize \mbox p}_1} }
\newcommand{\pedonez}{\bar{z}_{{\scriptsize \mbox p}} }
\newcommand{\pedendx}{x'_{{\scriptsize \mbox p}_N} }
\newcommand{\pedendy}{y'_{{\scriptsize \mbox p}_N} }
\newcommand{\pedendz}{\bar{z}'_{{\scriptsize \mbox p}} }

\newcommand{\pedend}{{\mathbf x}'_{{\scriptsize \mbox p}_{N}}}

\newcommand{\pworld}{{\mathbf x}_{{\scriptsize \mbox w}_{i}}}
\newcommand{\pworldx}{x_{{\scriptsize \mbox w}_i} }
\newcommand{\pworldy}{y_{{\scriptsize \mbox w}_i} }
\newcommand{\pworldz}{z_{{\scriptsize \mbox w}_i} }

\newcommand{\pcam}{{\mathbf x}_{{\scriptsize \mbox c}_i} }
\newcommand{\pcamx}{x_{{\scriptsize \mbox c}_i} }
\newcommand{\pcamy}{y_{{\scriptsize \mbox c}_i} }
\newcommand{\pcamz}{z_{{\scriptsize \mbox c}_i} }

\newcommand{\forcealpha}{{\mathbf f}_\alpha}
\newcommand{\desiredvelocity}{v_\alpha^0}
\newcommand{\currentdirection}{{\mathbf e}_\alpha}
\newcommand{\currentvelocity}{{\mathbf v}_\alpha}
\newcommand{\relaxation}{\tau_\alpha}

\newcommand{\repulsiveped}{{\mathbf f}_{\alpha \beta}}
\newcommand{\repulsivewall}{{\mathbf f}_{\alpha i}}

\newcommand{\repulsivestrength}{a_\alpha}
\newcommand{\repulsiverange}{b_\alpha}
\newcommand{\peddistance}{{\mathbf d}_{\alpha \beta}}

\newcommand{\ellipticalhelbing}{w_{\alpha \beta}}

\newcommand{\movementdirection}{{\mathbf n}_\alpha}
\newcommand{\perpendiculardirection}{{\mathbf p}_\alpha}
\newcommand{\vrel}{v_{\mathrm{rel}}}

\newcommand{\trajectoryalphaprime}{{{\cal T}'_{\alpha}}}
\newcommand{\trajectoryalpha}{{{\cal T}_\alpha}}

\newcommand{\point}{\mathbf x}

\newcommand{\euler}{e}

\newcommand{\alphai}{{{\scriptsize \mbox {$\alpha$} }_i}}

\newcommand{\TS}{\rule{0pt}{2.5ex}}       
\newcommand{\BS}{\rule[-1.1ex]{0pt}{0pt}} 

\journal{Transportation Research C}

\begin{document}

\begin{frontmatter}



\title{Kinects and Human Kinetics: A New Approach for Studying Crowd Behavior}


\author[1,2]{Stefan Seer\corref{cor1}}
\ead{stefan.seer@ait.ac.at, seer@mit.edu}
\cortext[cor1]{Corresponding Author}
\address[1]{Austrian Institute of Technology (AIT), Giefinggasse 2, 1210 Vienna, Austria}

\author[1]{Norbert Br\"andle}
\ead{norbert.braendle@ait.ac.at}

\author[2]{Carlo Ratti}
\ead{ratti@mit.edu}
\address[2]{MIT Senseable City Lab, Massachusetts Institute of Technology (MIT), 77 Massachusetts Avenue, 02139 Cambridge, MA, USA}

\begin{abstract}
Modeling crowd behavior relies on accurate data of pedestrian movements at a high level of detail. Imaging sensors such as cameras provide a good basis for capturing such detailed pedestrian motion data. However, currently available computer vision technologies, when applied to conventional video footage, still cannot automatically unveil accurate motions of groups of people or crowds from the image sequences. We present a novel data collection approach for studying crowd behavior which uses the increasingly popular low-cost sensor Microsoft Kinect. The Kinect captures both standard camera data and a three-dimensional depth map. Our human detection and tracking algorithm is based on agglomerative clustering of depth data captured from an elevated view -- in contrast to the lateral view used for gesture recognition in Kinect gaming applications. Our approach transforms local Kinect 3D data to a common world coordinate system in order to stitch together human trajectories from multiple Kinects, which allows for a scalable and flexible capturing area.       
At a testbed with real-world pedestrian traffic we demonstrate that our approach can provide accurate trajectories from three Kinects with a Pedestrian Detection Rate of up to 94\% and a Multiple Object Tracking Precision of 4~cm. Using a comprehensive dataset of 2240 captured human trajectories we calibrate three variations of the Social Force model. The results of our model validations indicate their particular ability to reproduce the observed crowd behavior in microscopic simulations.
\end{abstract}


\end{frontmatter}



\input{introduction}
\input{tracking}
\input{evaluation}
\input{results}
\input{conclusion}

\section{Acknowledgments}
We would like to thank Jim Harrington, of the MIT School of Architecture and Planning, and Christopher B. Dewart, of the MIT Department of Architecture, for their support in the installation of equipment in the MIT's Infinite Corridor. The authors would also like to thank David Lee for his research assistance in designing and running the experiments.
Support is gratefully acknowledged from the MIT SMART program, the MIT CCES program, Audi-Volkswagen, BBVA, Ericsson, Ferrovial, GE and all the members of the Senseable City Consortium.

\bibliographystyle{model2-names}
\bibliography{references}







\end{document}

%% file: introduction.tex
\section{Introduction}
\label{introduction}
With 60\% of the world's population projected to live in urban areas by 2030, crowd management and modeling is becoming an urgent issue of global concern. A better understanding of pedestrian movement can lead to an improved use of public spaces, to the appropriate dimensioning of urban infrastructure (such as airports, stations and commercial centers), and, most importantly, to the design of cities that are more responsive to people and to that very fundamental human activity -- walking.
 
At the urban block and building model scale, predictions on crowd movement are usually being investigated using microscopic pedestrian simulation models. The development and calibration of such models requires highly accurate data on pedestrian movements. This data is provided by individual movement trajectories in space. 
Modeling human interaction behavior calls for an analysis of {\em all} people in a given scene.
At the same time, collecting large amounts of quantitative data on how people move in different environments is a very time consuming and elaborate process.

Traditionally, such data is collected by manually annotating the positions of people in individual  frames of recorded video data of  highly frequented areas (\cite{Antonini2006}; \cite{Berrou2005}). Sometimes, additional attributes such as age or gender are assigned during the annotation process. But manual annotation is particularly complex in dense scenes which limits the amount of data that can be analyzed.
As a result, large scale data on human motion can only be obtained from video, using tools for automatic vision-based detection and tracking of pedestrians.  Currently available computer vision methods suffer from several limitations, such as occlusions of static and moving objects, changing lighting conditions and background variations. For example, \cite{Breitenstein2011} describes a tracking approach that relies on two dimensional image information from a single, uncalibrated camera, without any additional scene knowledge. While this method shows an improved performance compared to other state-of-the-art results, occlusions at higher densities of people lead to missing detections or switching individuals.

In order to avoid severe occlusions most of today's commercially available people counter solutions use overhead sensors.
Due to their restricted view, multiple sensors are required for a larger capturing area. These sensors are often very expensive, so the observation of pedestrian movement on a larger spatial scale imposes high costs. Furthermore, commercial solutions usually do not provide access to trajectory data.

Semi-automated video approaches are described in \cite{Plaue2011} and \cite{Johansson2008}. They are based on the manual annotation of objects (e.g. people's heads) within very few images, which are then provided as input to an algorithm that tracks across different frames. While such systems have clear advantages in analyzing simple, low density scenes, they suffer from both high manual effort and less robust automatic tracking in complex scenarios.

Experimental setups represent another approach for collecting trajectories of individuals. In these setups participants can be equipped with distinctive wear such as colored hats for better identification. External factors such as lighting conditions can be controlled. As a result, the automated extraction of trajectories can be very robust. The free software  \emph{PeTrack} presented in \cite{Boltes2008} has been applied on video recordings of a bottleneck experiment. The automatic tracking approaches of \cite{Hoogendoorn2003a} and \cite{Hoogendoorn2005} collected trajectory data in a narrow bottleneck and a four-directional crossing flow experiment. Controlled experiments allow the setting of environmental conditions that are hard to observe in real world circumstances, as in \cite{Daamen2012a} where emergency settings were reenacted including acoustic and visual signals. However, these setups only allow for a limited sample size and include a significant bias in the data since participants are usually aware of being observed.
 
In this paper we propose a novel approach for automatically collecting highly accurate and comprehensive data on individual pedestrian movement using the Microsoft Kinect -- a motion sensing input device which was originally developed for the Xbox 360 video game console (\cite{MicrosoftKinect}). Our proposed use of the Kinect represents a very economical way to collect movement data which overcomes many of the above described limitations. Furthermore, thanks to the increased richness of the sensed data in three dimensions, it can open the way to more sophisticated, fine grain analyses of crowd movement.
 
\begin{figure}[htb]
  \centering
  \begin{minipage}[c]{0.45\textwidth}
    \centering
    \includegraphics[height=5cm]{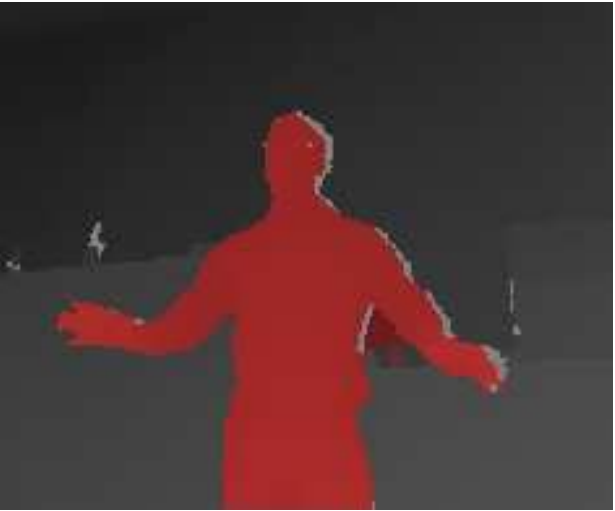}
  \\(a)\\
  \end{minipage}
  \hspace{0.5cm}
  \begin{minipage}[c]{0.45\textwidth}
   \centering
    \includegraphics[height=5cm]{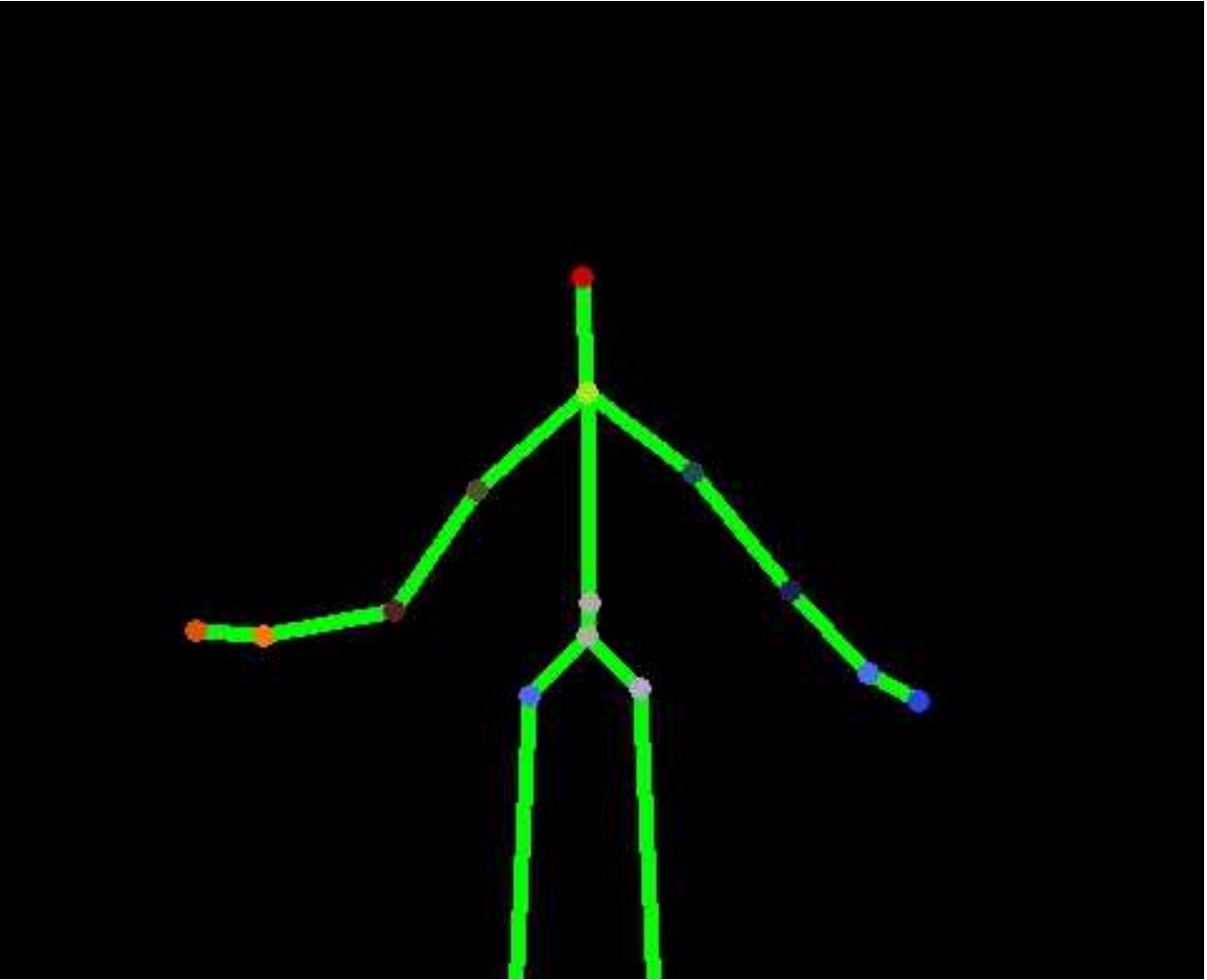}
   \\(b)\\
  \end{minipage}
  \caption{Microsoft Kinect provides the depth data stream (a), with the detected person in red and different gray levels encoding the depth information, and skeletal tracking (b).}
  \label{fig:KinectDepthSkeletal}
\end{figure}

The Kinect is an inexpensive sensor that delivers not only camera information, but also a 3D depth map which is particularly useful for computer vision and pattern recognition purposes. The Kinect was originally designed to accurately detect three dimensional positions of body joints (\cite{Shotton2011}) and to estimate human pose (\cite{Girshick2011}). Figure~\ref{fig:KinectDepthSkeletal} illustrates the \emph{skeletal tracking} which is the key component of the video game user interface. With its built-in functionality, the Kinect can detect up to six people (two of them using the skeletal tracking) provided that all persons face the sensor in frontal view with their upper bodies visible. Since its market introduction in 2010, the Kinect has also been used in a broad variety of other research fields: \cite{Noonan2011} showed the use of the Kinect for tracking body motions in clinical scanning procedures. Animation of the hand avatar in a virtual reality setting by combining the Kinect with wearable haptic devices was developed in \cite{Frati2011}. The Kinect was used in \cite{Izadi2011} to create detailed three dimensional reconstructions of an indoor scene. \cite{Weiss2011} presented a method for human shape reconstruction using three dimensional and RGB data provided by the Kinect. 

To the best of our knowledge, the Microsoft Kinect has not yet been used to obtain data for modeling crowd behavior.  For its purpose as a user interface, the Kinect has an implemented capability for skeletal tracking of individuals. However, this feature cannot be directly used for measuring crowd movement. Given the necessary conditions for its built-in people detector, a single Kinect is not able to deliver stable detections of all individuals in crowded scenes with more than six people and mutual occlusions severely affect the detection performance.
 
\begin{figure}[htb]
  \centering
  \begin{minipage}[c]{0.60\textwidth}
    \centering
  \includegraphics[height=5.5cm]{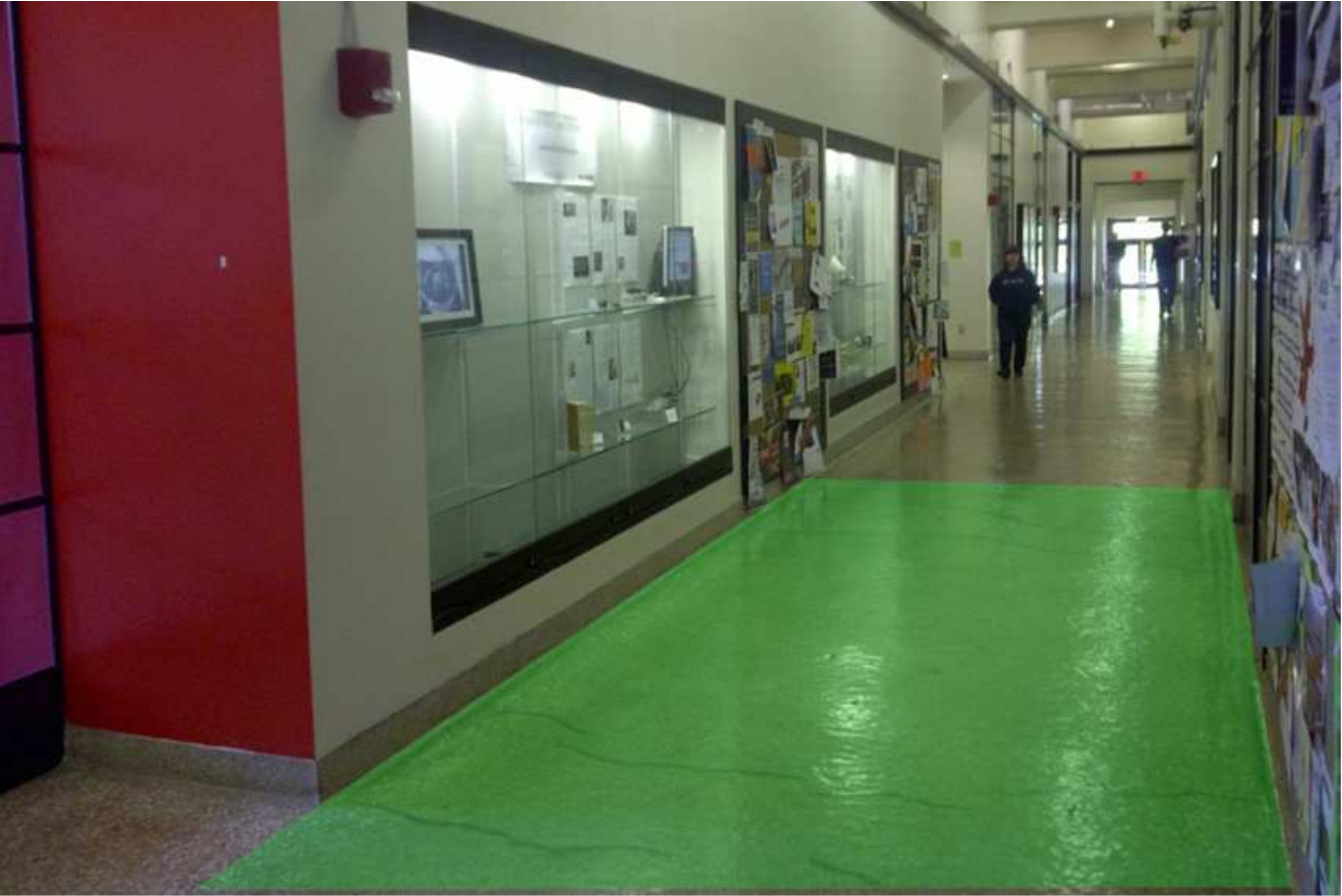}
  \\(a)\\
  \end{minipage}
  \hspace{0.5cm}
  \begin{minipage}[c]{0.30\textwidth}
   \centering
  \includegraphics[height=5.5cm]{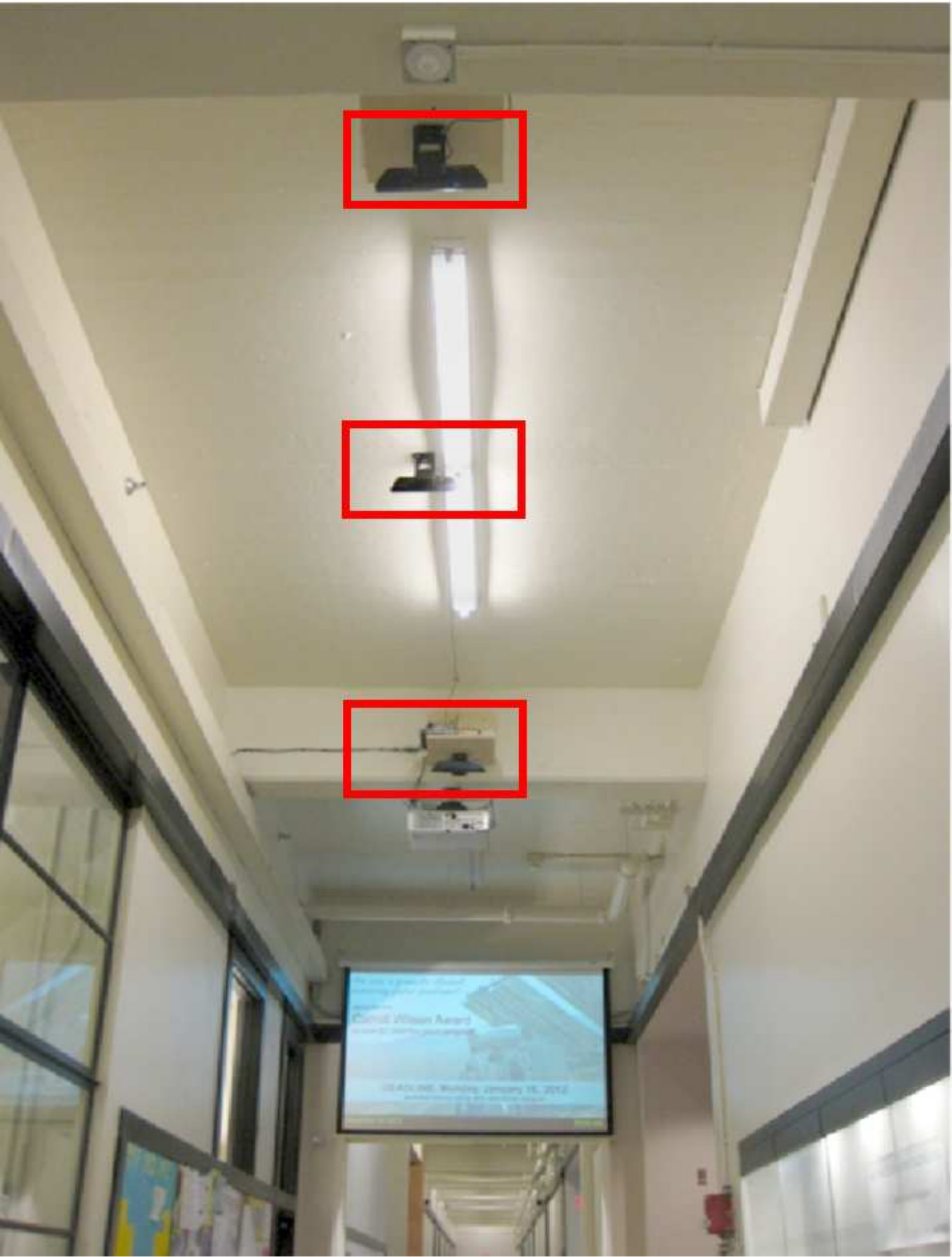}
   \\(b)\\
  \end{minipage}
  \caption{MIT's Infinite Corridor with (a) the observed area (green) and (b) the Kinect setting on the ceiling.}
  \label{fig:InfiniteCorridor}
\end{figure}

In this paper, we demonstrate how the Microsoft Kinect can be used to obtain tracking data for crowd modeling. It has the potential to become an invaluable tool for analyzing pedestrian movement, overcoming most of the limitations of hitherto used automated and semi-automated capturing systems. Furthermore, we aim to make the following detailed contributions:
\begin{enumerate}
\item We present algorithms for processing depth data from multiple Kinects to retrieve pedestrian trajectories from an elevated view.
\item We demonstrate the performance of our algorithms in a real world setup, also addressing the setting details and sensor calibration.
\item We use an extensive data set derived from our approach to calibrate and compare state-of-the-art microscopic pedestrian simulation models.
\end{enumerate}

We combined three Kinect sensors and collected a large dataset on crowd movement inside the Massachusetts Institute of Technology (MIT)'s Infinite Corridor, the longest hallway that serves as the most direct indoor route between the east and west ends of the campus and is highly frequented by students and visitors. 
Figure~\ref{fig:InfiniteCorridor}a shows the area identified for the data collection in this work, and Figure~\ref{fig:InfiniteCorridor}b shows the Kinect sensors mounted at the ceiling. In order to observe various pedestrian behaviors we performed different walking experiments.

This paper is structured as follows: Section~\ref{tracking} outlines the setting for measuring human motion data using the Kinect. We also explain the calibration process needed in order to derive world coordinate data from the Kinect sensors. Furthermore, we describe the algorithms for detecting and tracking of humans using multiple Kinects. Section~\ref{evaluation} provides evaluation results showing the tracking performance in a setting with single and multiple Kinects. Section~\ref{results} describes the walking experiments and data collection at MIT's Infinite Corridor. We describe how these data sets can be used for the calibration of crowd models and provide results from the calibration and validation of three simulation models. Section~\ref{conclusion} concludes the results and gives an outlook for further research.

%% file: tracking.tex
\section{Human Detection and Tracking}
\label{tracking}
\begin{figure}[htb]
  \centering
  \begin{minipage}[c]{0.3\textwidth}
    \centering
    \includegraphics[scale=0.4]{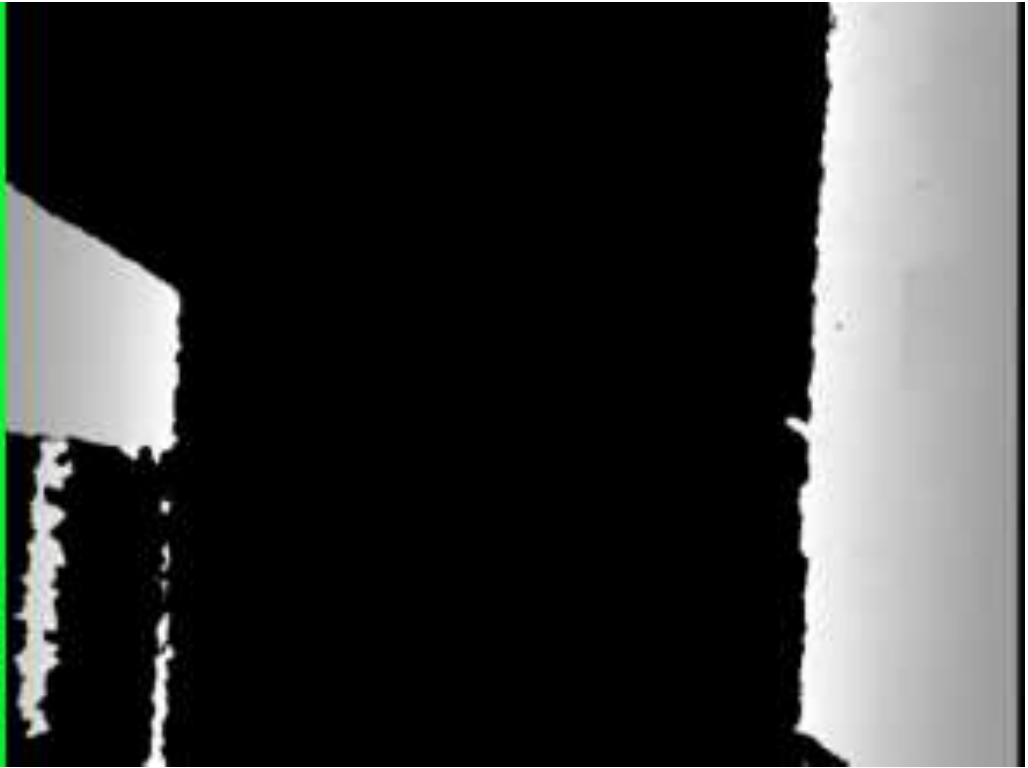}
  \end{minipage}
  \hspace{0.1cm}
  \begin{minipage}[c]{0.3\textwidth}
   \centering
    \includegraphics[scale=0.4]{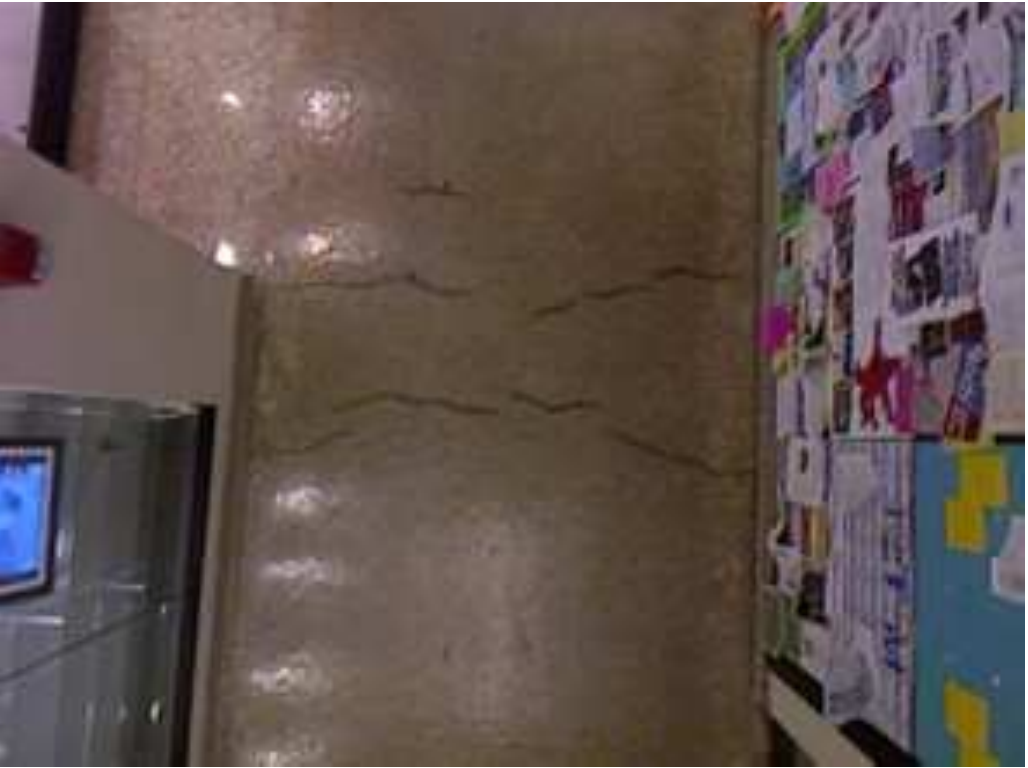}
  \end{minipage}
  \hspace{0.1cm}
  \begin{minipage}[c]{0.3\textwidth}
   \centering
    \includegraphics[scale=0.4]{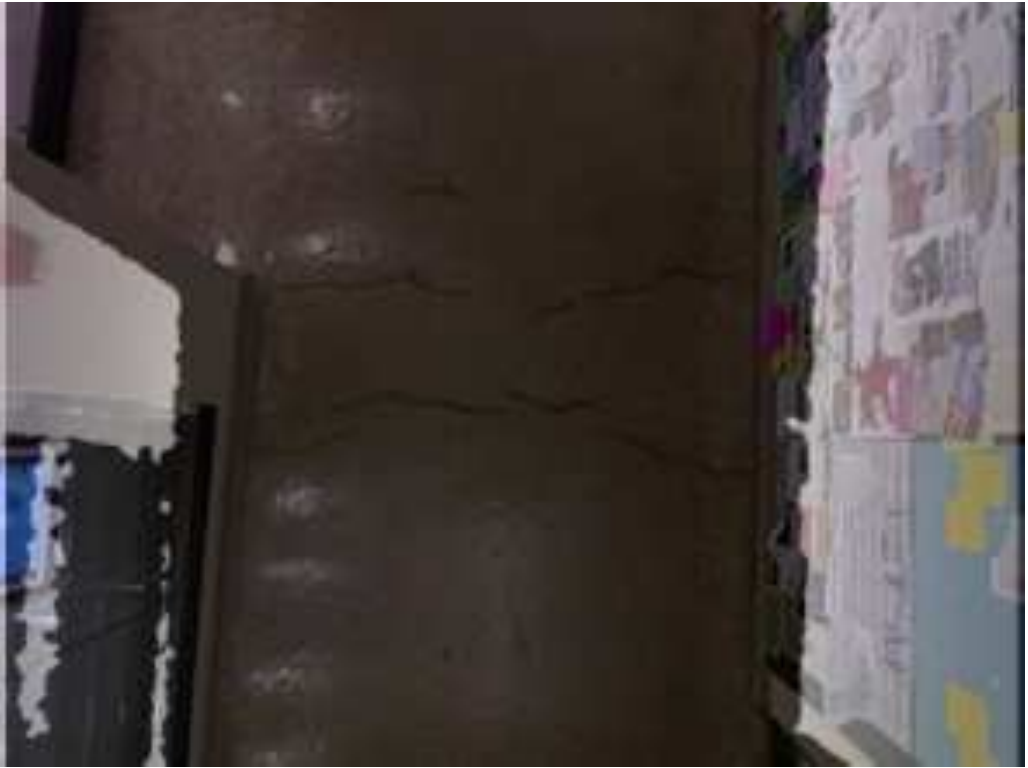}
  \end{minipage}
  \vspace{0.15cm}
    \\Sensor $S_1$\\
  \vspace{0.3cm}
 \begin{minipage}[c]{0.3\textwidth}
    \centering
    \includegraphics[scale=0.4]{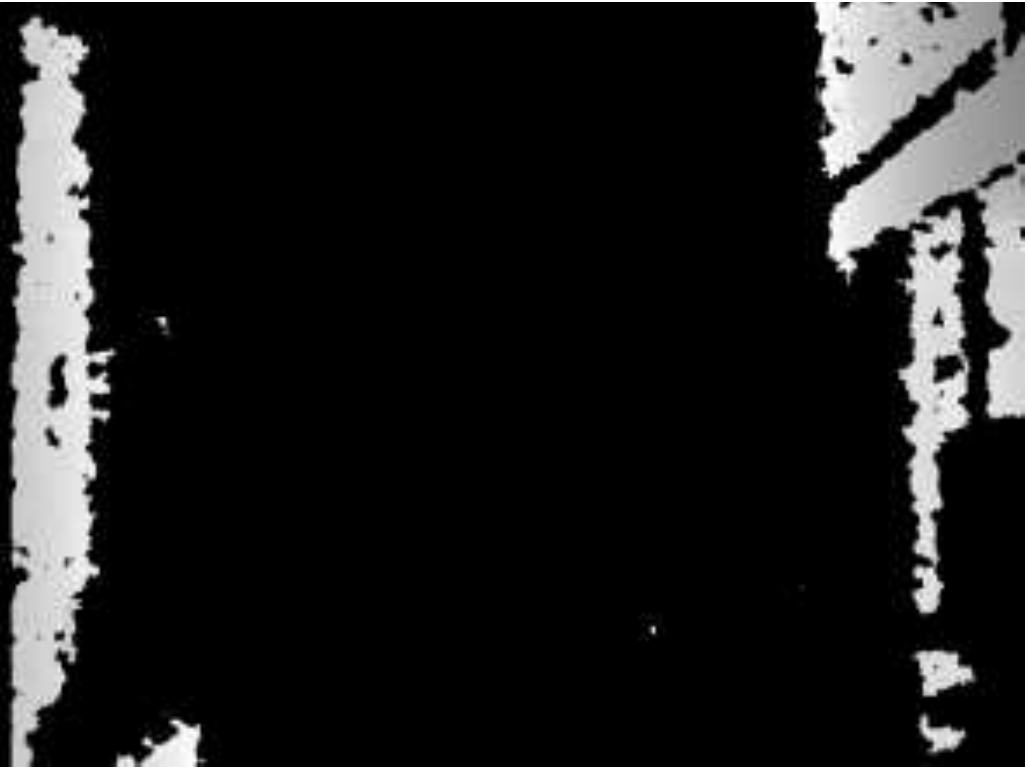}
  \end{minipage}
  \hspace{0.1cm}
  \begin{minipage}[c]{0.3\textwidth}
   \centering
    \includegraphics[scale=0.4]{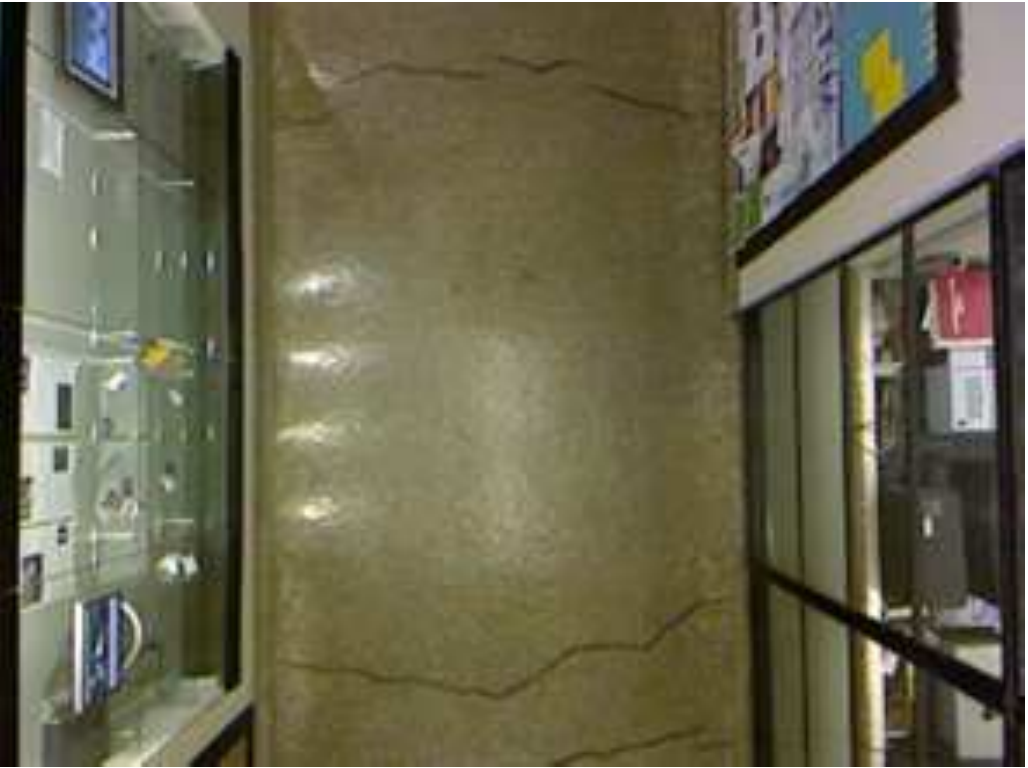}
  \end{minipage}
  \hspace{0.1cm}
  \begin{minipage}[c]{0.3\textwidth}
   \centering
    \includegraphics[scale=0.4]{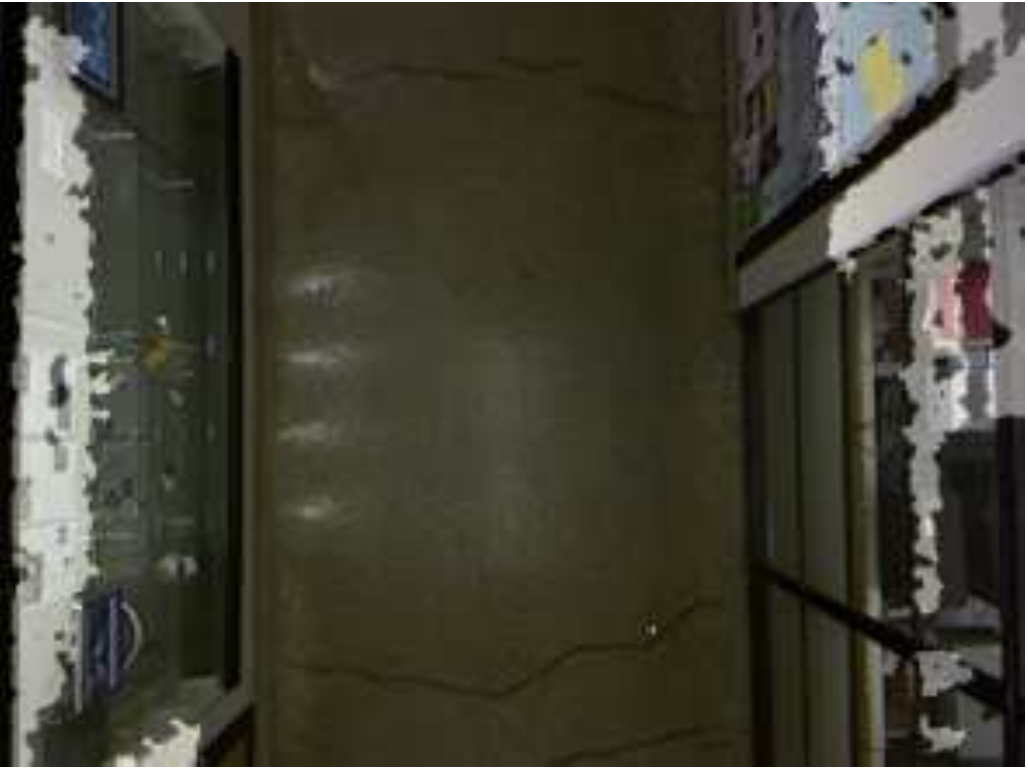}
  \end{minipage}
  \vspace{0.15cm}
    \\Sensor $S_2$\\
  \vspace{0.3cm}
  \begin{minipage}[c]{0.3\textwidth}
    \centering
    \includegraphics[scale=0.4]{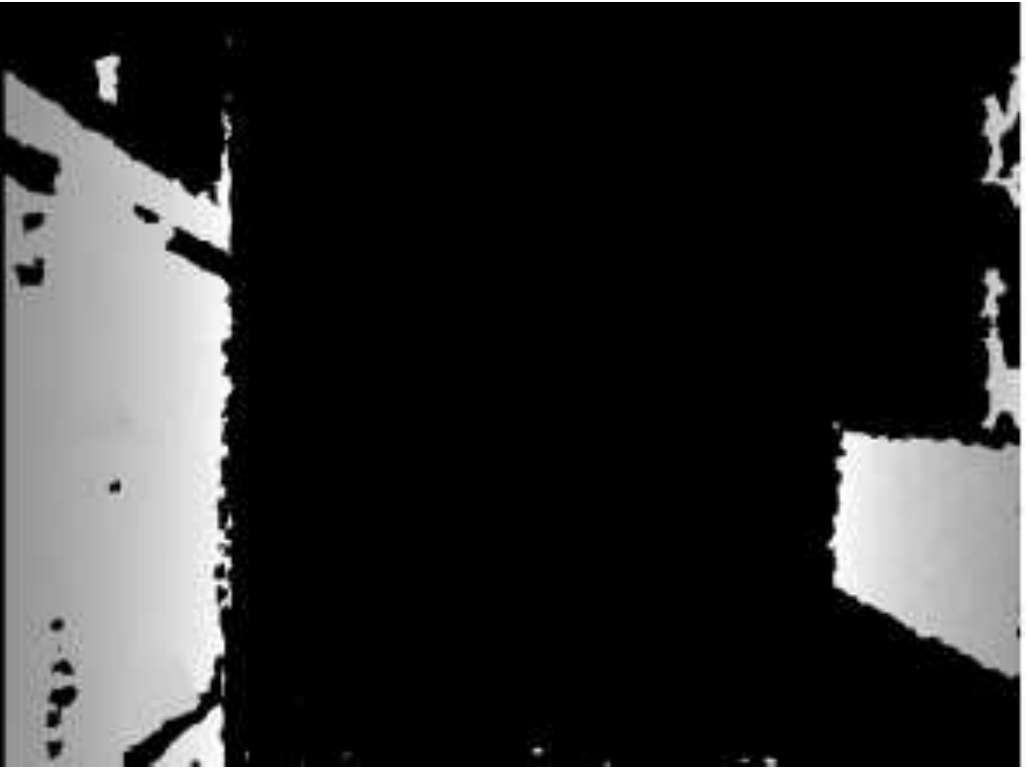}
  \end{minipage}
  \hspace{0.1cm}
  \begin{minipage}[c]{0.3\textwidth}
   \centering
    \includegraphics[scale=0.4]{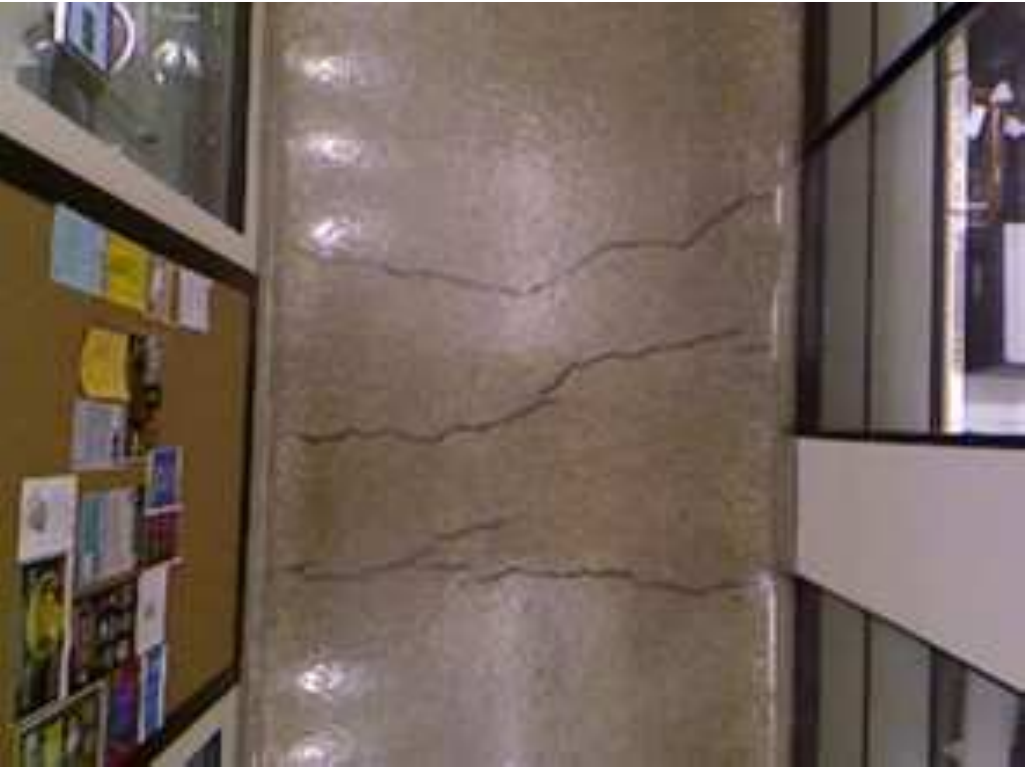}
  \end{minipage}
  \hspace{0.1cm}
  \begin{minipage}[c]{0.3\textwidth}
   \centering
    \includegraphics[scale=0.4]{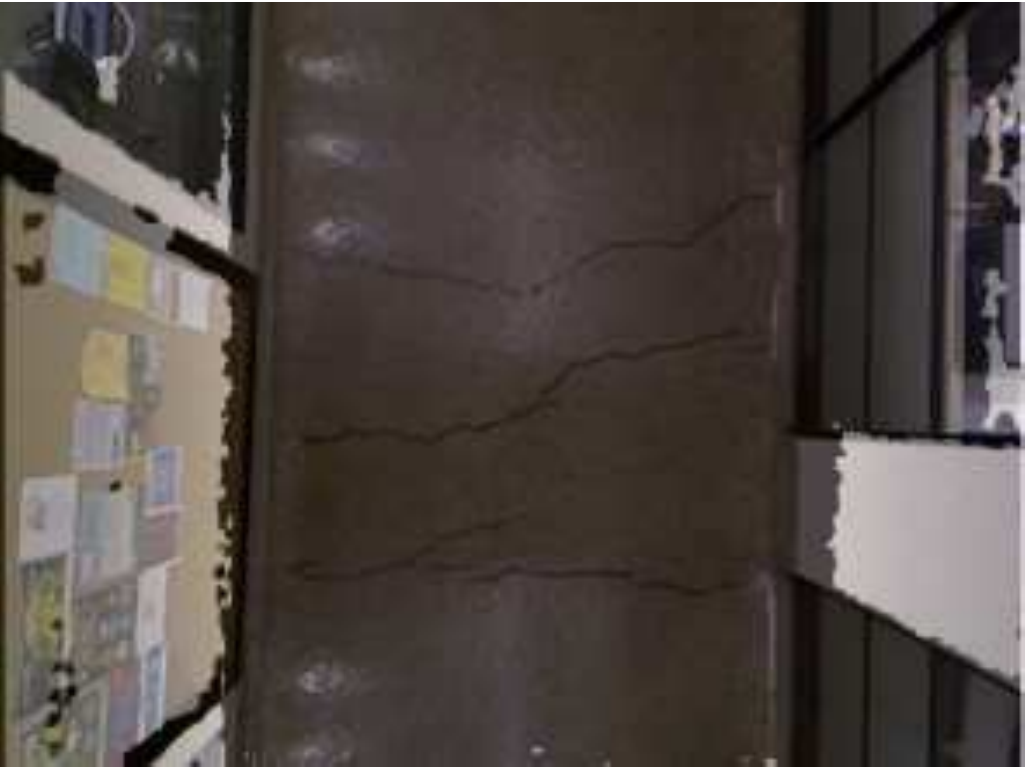}
  \end{minipage}
  \vspace{0.15cm}
    \\Sensor $S_3$\\
  \caption{Kinect sensor field of view; raw depth data stream (left), RGB stream (middle) and both data streams in an overlay (right).}
  \label{fig:SensorViews}
\end{figure}

Detailed knowledge of pedestrian flows is of vital importance for the calibration and validation of microscopic pedestrian simulation models.  The Kinect can be thought of as a modified camera. Like a traditional camera it provides a sequence of standard RGB color frames. In addition, it delivers a 3-dimensional depth image for each frame. The depth image of a scene tells us the distance of each point of that particular scene from the Kinect. Depth images and RGB color images are both accessible with the \emph{Kinect for Windows SDK} by \cite{KinectSDK}. Figure~\ref{fig:SensorViews} illustrates a snapshot of the depth image, the RGB image and a combination of depth and RGB from three Kinects mounted at a height of 4.5 meters and a top view position in the MIT's Infinite Corridor. With this setup a section of 6 meters of the corridor can be captured.  Note that the glass case introduces a significant amount of artifacts due to specular reflections.  In order to meet privacy concerns -- most of the observed persons are not aware of any data collection experiment -- our approach does not process RGB information from the visible spectrum. 

In order to compute pedestrian trajectories from depth image sequences of multiple Kinects, it is necessary to 1) map depth information from individual Kinect sequences into a common world coordinate system, 2) group depth information from a single Kinect in the world coordinate system into individual pedestrians and track the pedestrians to obtain trajectories and 3) stitch pedestrian trajectories from multiple Kinect sensors. These three steps are described in the following subsections.

\subsection{Obtaining World Coordinates}
\label{setting}

A Kinect sensor $\kinect$ from a set of $K$ devices generates a time series of $640\times480$ depth pixel images. Each depth image encodes a set of valid three-dimensional points  $\pcam = [\pcamx \; \pcamy \; \pcamz]^T$, with $i \leq 640\times480$, in the local Kinect 3D camera coordinate system, computed with the value of the focal length $f$ provided by \cite{KinectSDK}. 
The physical constraints of the Kinect 3D-measurement setup limit the range of $\pcamz$  within which reliable depth data can be computed  to a maximum distance of 4 meters. Objects which are located more than 4 meters away from the sensor can not be captured.    

A human trajectory $\trajectory$ is denoted as a sequence of $N$ four-dimensional vectors 
\begin{equation}
\trajectory  = \{[t_i \; \pworldx \; \pworldy \; \pworldz]^T\}_{i = 1  \ldots N,}  
\label{eq:trajectory}
\end{equation}
where the vectors are composed of a timestamp $t_i$ and a 3D position $\pworld = [\pworldx \; \pworldy \; \pworldz]^T$ in a common world coordinate system:   
For a trajectory to represent people walking throughout the sensing areas of multiple Kinect sensors, the points of the local 3D coordinate systems of the mounted Kinect sensors must first be mapped to the world coordinate system. 

The actual point mapping between the coordinate system of sensor $\kinect$ and the world coordinate system is represented by a rigid transformation, composed of a translation vector $\translation_k$ between the two origins of the coordinate systems and 
a $3 \times 3$ rotation matrix $\rotation_k$ such that

\begin{equation}
\pworld = \rotation_k\pcam + \translation_k .
\label{eq:rigid_transform}
\end{equation}
The three parameter values for translation $\translation_k$ of sensor $\kinect$ and its three rotation angles in  $\rotation_k$ are determined by a set of $M$ point matches~$<\pworld, \pcam>$, $i \in M$ and subsequently minimizing the error 
\begin{equation}
E = \sum_{i=1}^{M}\left | \pworld - \rotation_k \pcam  - \translation_k \right|^2
\label{eq:rigid_transform_error}
\end{equation}
by solving an overdetermined equation system as described in \cite{Forsyth2002}. 

\begin{figure}[p]
  \centering
  \begin{minipage}[c]{0.45\textwidth}
    \centering
    \includegraphics[scale=0.38]{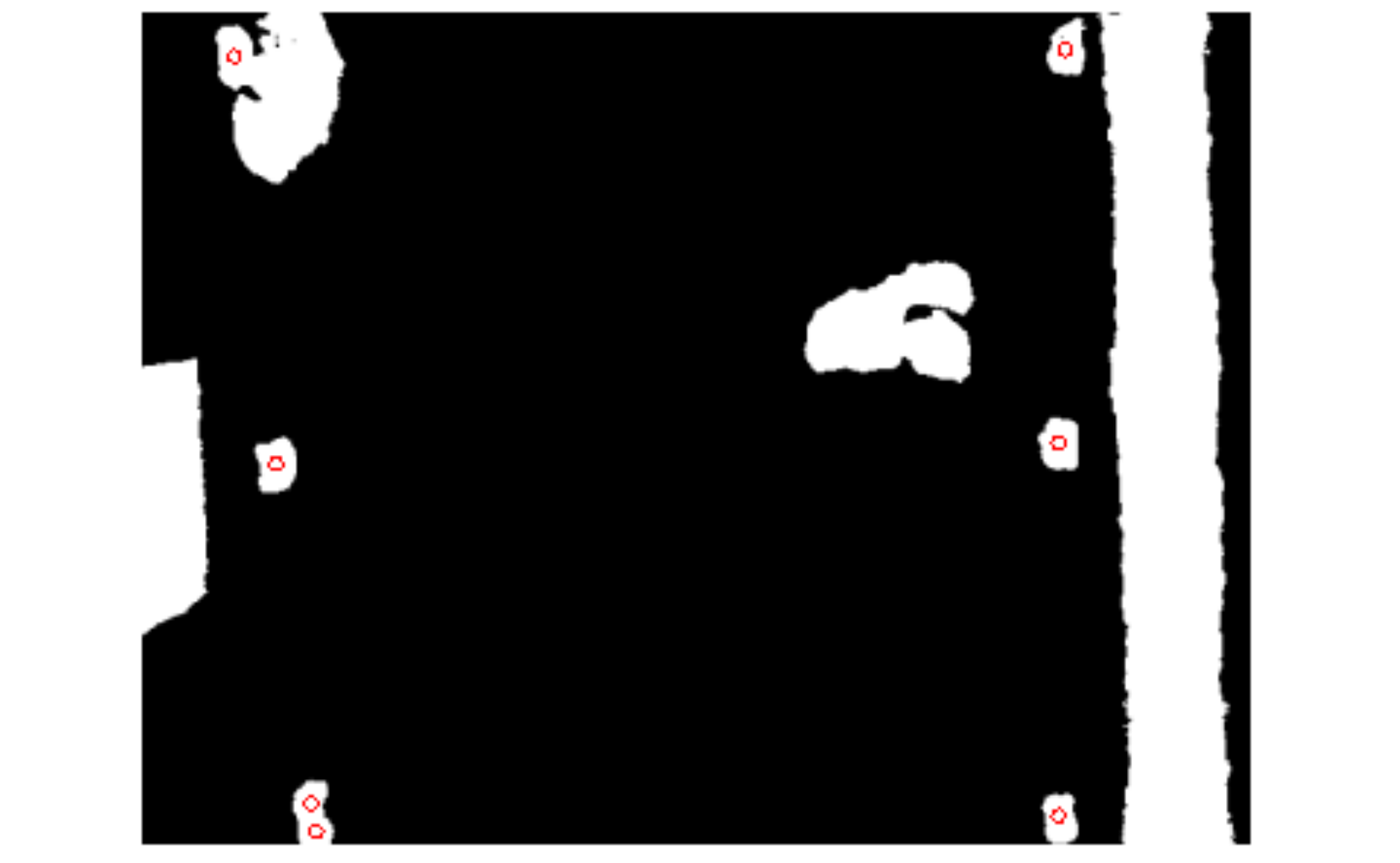}
  \end{minipage}
  \hspace{0.5cm}
  \begin{minipage}[c]{0.45\textwidth}
   \centering
    \includegraphics[scale=0.4]{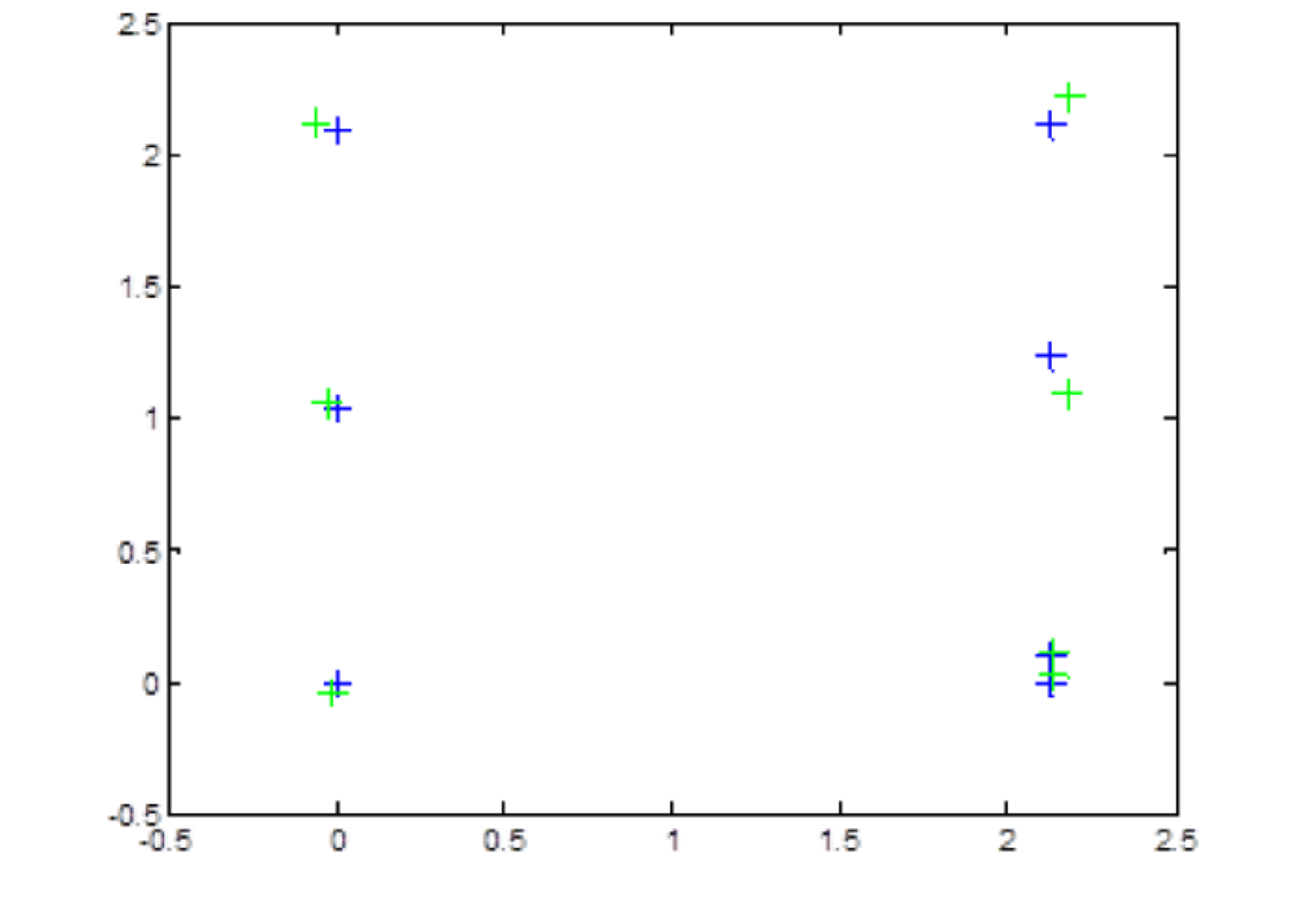}
  \end{minipage}
  \vspace{0.15cm}
    \\Sensor $S_1$\\
  \vspace{0.3cm}
  \begin{minipage}[c]{0.45\textwidth}
    \centering
    \includegraphics[scale=0.38]{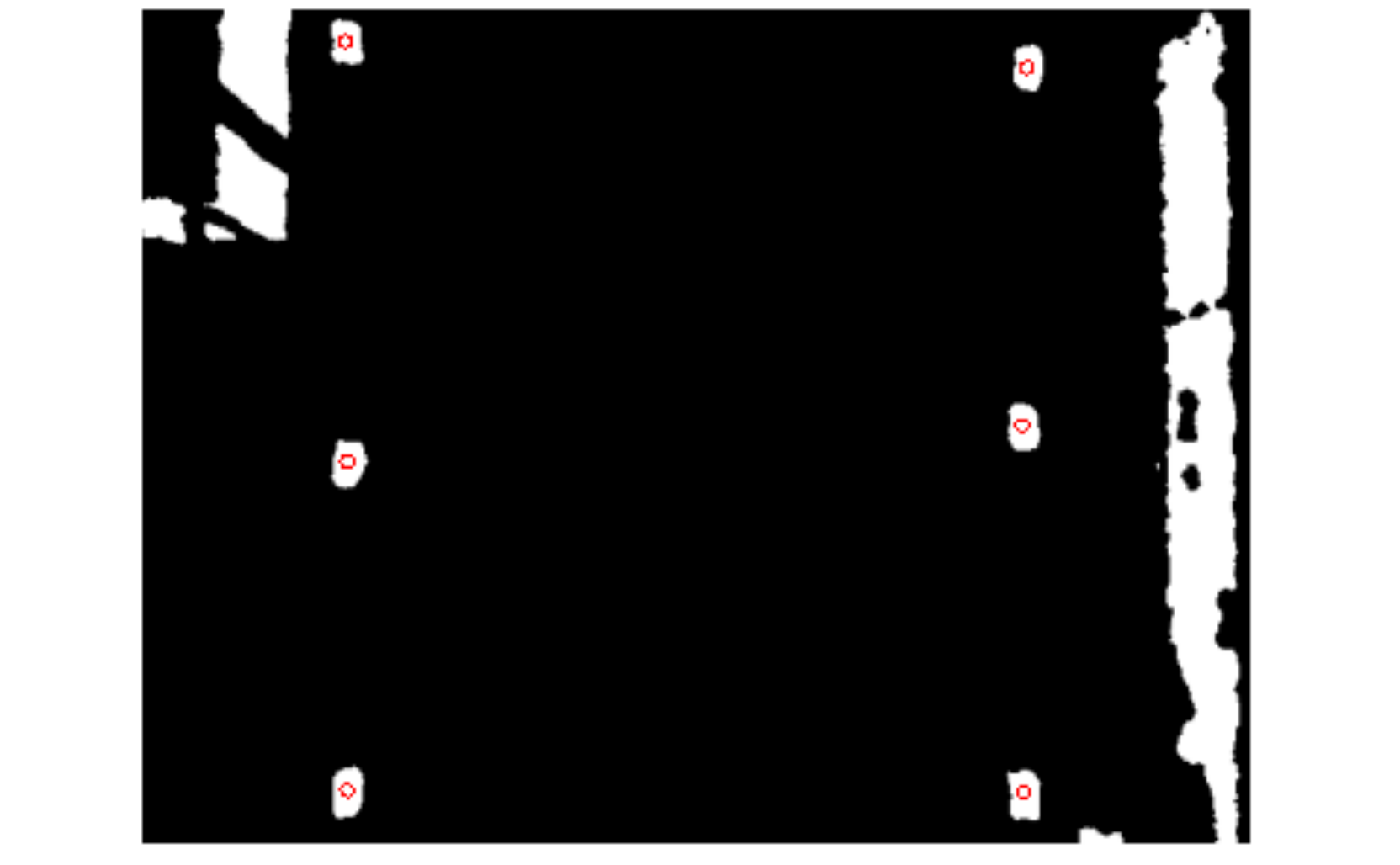}
  \end{minipage}
  \hspace{0.5cm}
  \begin{minipage}[c]{0.45\textwidth}
   \centering
    \includegraphics[scale=0.4]{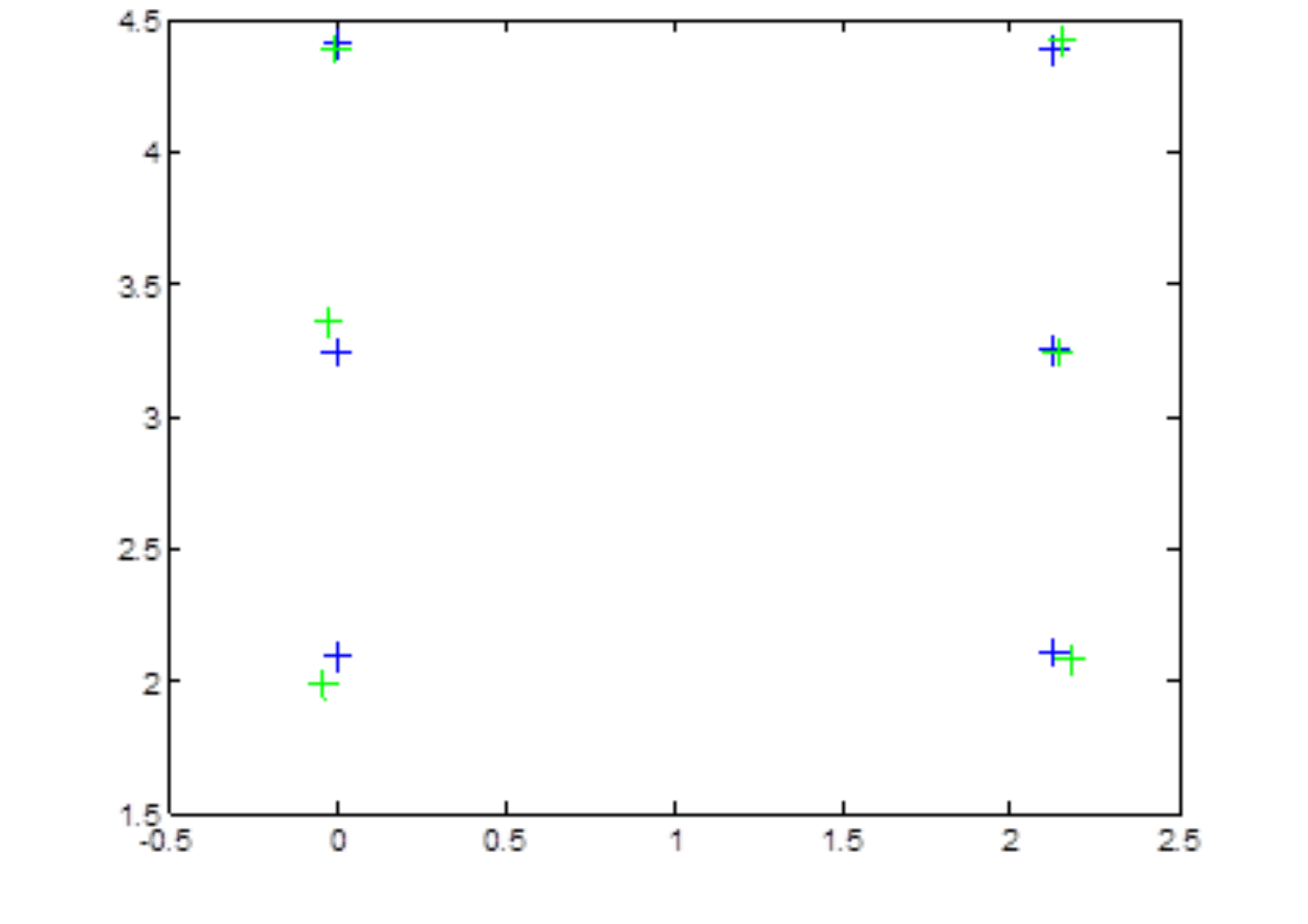}
  \end{minipage}
  \vspace{0.15cm}
    \\Sensor $S_2$\\
  \vspace{0.3cm}
  \begin{minipage}[c]{0.45\textwidth}
    \centering
    \includegraphics[scale=0.38]{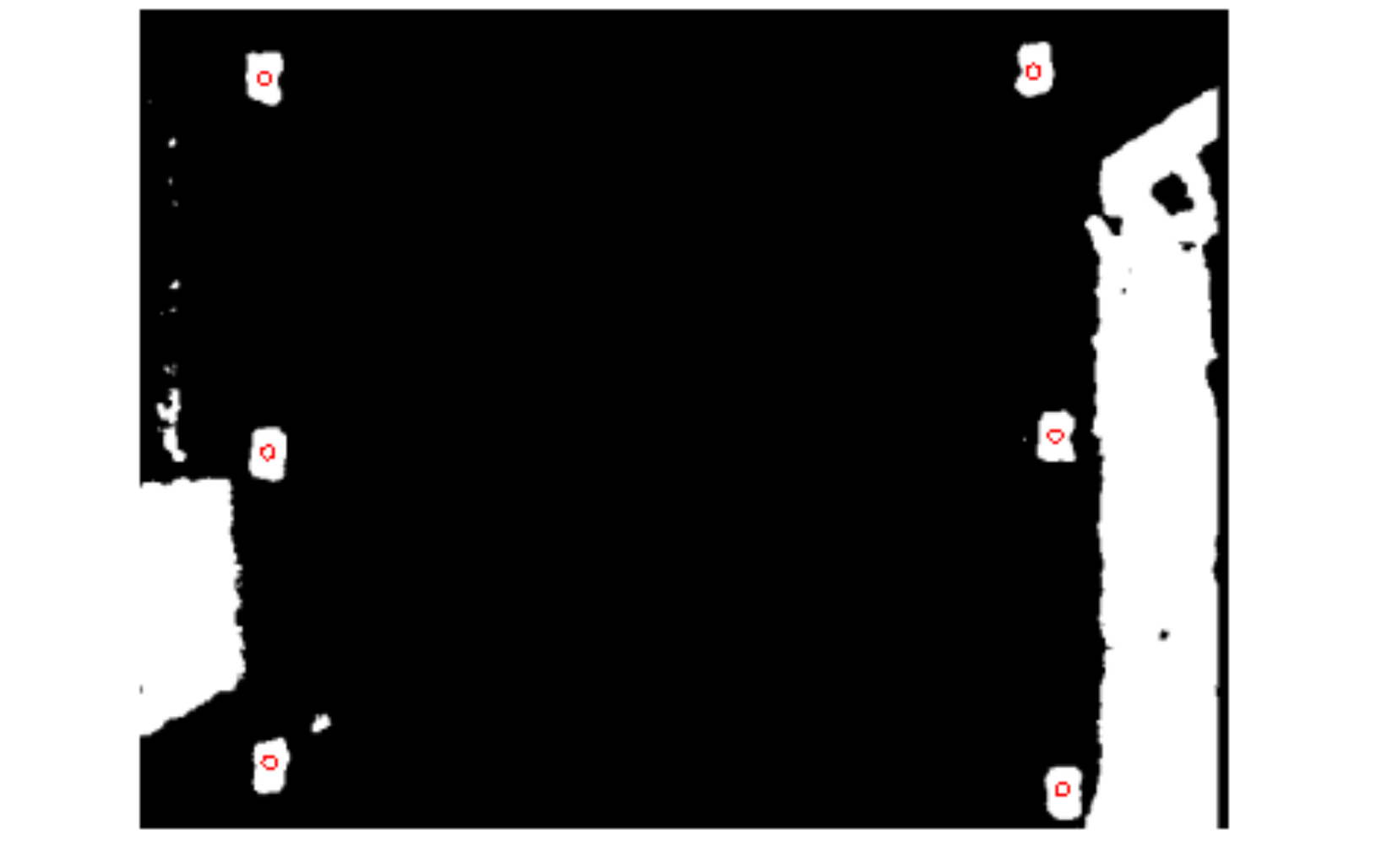}
  \end{minipage}
  \hspace{0.5cm}
  \begin{minipage}[c]{0.45\textwidth}
   \centering
    \includegraphics[scale=0.4]{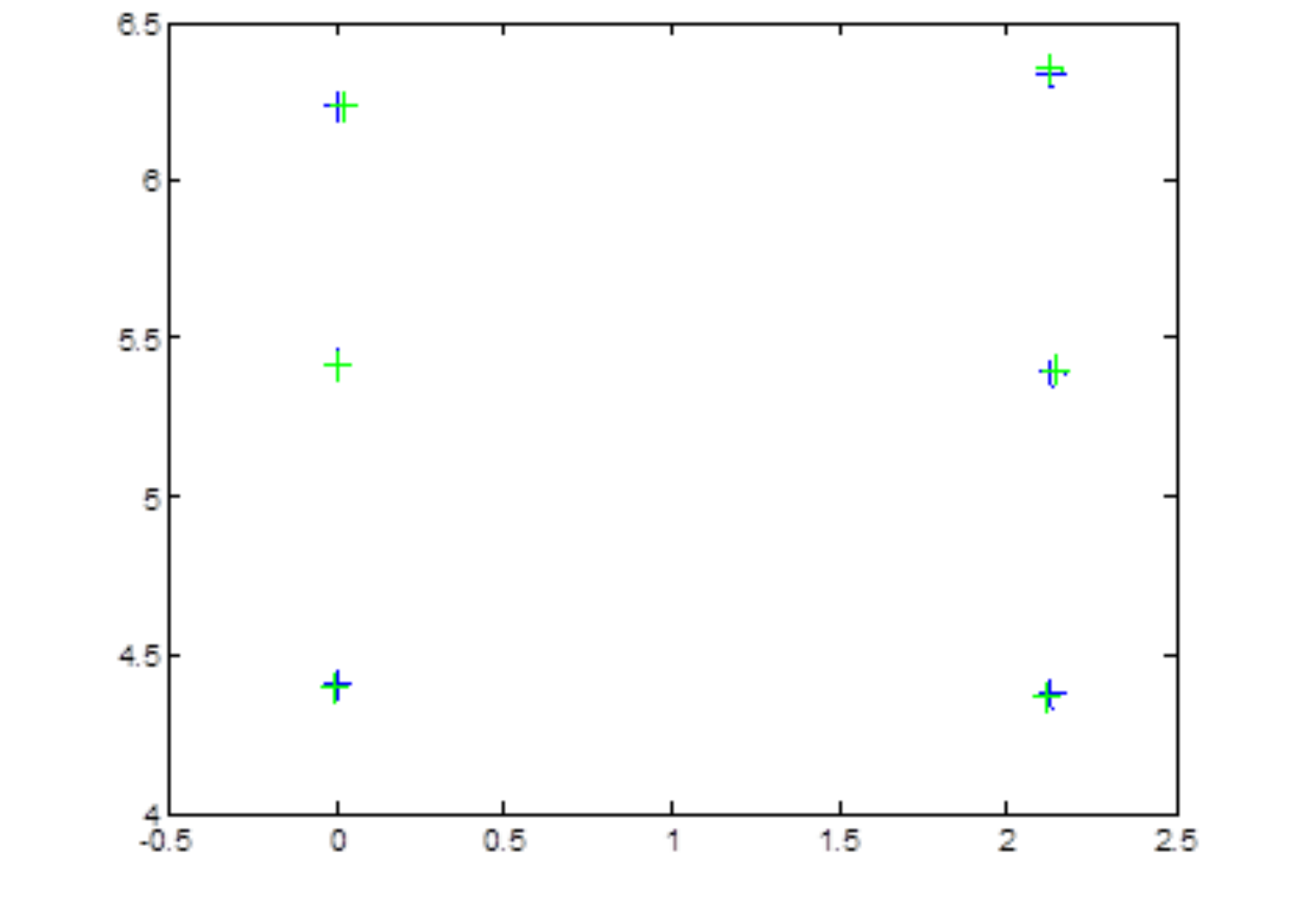}
  \end{minipage}
  \vspace{0.15cm}
    \\Sensor $S_3$\\
  \caption{Left column - Raw data from sensor including reference points (red circles); Right column - sensor calibration results with measured reference (blue) and estimated (green) points.}
  \label{fig:CalibrationSensorViews}
\end{figure}

We determine the $M$ point matches~$<\pworld, \pcam>$ in world coordinates $\pworld$ manually from the depth images. Since only depth information and no visual information is available for the sensed area, sensor calibration must be based on pre-determined calibration objects with well-defined depth discontinuities. Our sensor calibration setup is composed of a rectangular piece of cardboard placed on a tripod. The reference points in world coordinates $\pworld$ are determined as the center of gravity of the extracted cardboard corners in the depth images. The raw depth data including the reference points and the results of the calibration for all sensors are shown in Figure~\ref{fig:CalibrationSensorViews}.
Table~\ref{tb:SensorCalibrationPerformance} shows that the Root-Mean-Square Error (RMSE) between the reference points in the world coordinates and the reference points in camera coordinates transformed with \eqref{eq:rigid_transform} lies within the range of a few centimeters.
\begin{table}[h]
  \centering
  \begin{tabular}{  l | c | c | c |}
    \cline{2-4}
    & Sensor $S_1$ & Sensor $S_2$ & Sensor $S_3$ \TS\BS \\ \hline
    \multicolumn{1}{|c|}{RMSE} & 64 mm & 67 mm & 19 mm  \TS\BS \\ \hline
  \end{tabular}
\caption{Accuracy of calibration computed on reference points.}
\label{tb:SensorCalibrationPerformance}
\end{table}

\subsection{Detection and Tracking Algorithm}
\label{pedestrianDetection}

Let $\depthpointset$ denote the set of points $\pworld$ obtained by applying the rigid transform~\eqref{eq:rigid_transform} 
to the 3D camera coordinates from a Kinect depth image. The objective of human detection is to extract from $\depthpointset$ connected sets of points belonging to a person and to represent the person with a  
point $\ped$. Human tracking associates detections of individuals over time.
Human detection is composed of the following steps:

\begin{enumerate}
\item {\bf Data Reduction by Background Subtraction}. Identifying a set of points which do not change or only change slowly over time -- the background -- supports the segmentation of walking persons from other objects and reduces the number of depth points to be processed. This can be achieved by classic background subtraction techniques from the domain of video analysis, e.g. the adaptive background modeling with Gaussian Mixture Models described in \cite{Stauffer2000}. In our particular case of the Infinite Corridor, the background model is handcrafted, since the locations of background objects such as walls are well-known in advance. 

\item {\bf Data Reduction by Cutoff.} The cutoff step first removes all 3D points which remain after background subtraction with height $\pworldz$ larger than a tall person's height, e.g. 2.1~meters for adults, and all 3D points with height $\pworldz$ smaller than a typical upper body region, e.g. 1.5~meters. The second cutoff value determines the minimal height of detectable persons, and is necessary to exclude noisy measurements of objects near the floor. Applying the cutoff values to  $\pworldz$ results in a subset $\depthpointsetprime$.  

\item {\bf Hierarchical Clustering on the Reduced Set.} In order to group the points $\depthpointsetprime$ into natural clusters corresponding to individual persons, we first build a cluster tree by agglomerative clustering with the complete-linkage algorithm (\cite{DudaandHart}). For computational reasons we randomly select a subset $\depthpointsetdprime$ of $R$ points from $\depthpointsetprime$ for clustering, where typically $R=500$.  The complete-linkage algorithm uses the following distance $d(\clusteridprime,\clusterjdprime)$ to measure the dissimilarity between subsets of $\depthpointsetdprime$: 
\begin{equation}
d(\clusteridprime,\clusterjdprime) = \max_{\substack{ \mathbf{x} \in \clusteridprime \\ \mathbf{x'} \in \clusterjdprime }} 
||\mathbf{x} - \mathbf{x}' ||,   
\label{eq:dmax}
\end{equation}
with $|| \cdot ||$ as the Euclidean distance. Using metric \eqref{eq:dmax} avoids elongated clusters and is advantageous when the true clusters are compact and roughly equal in size (\cite{DudaandHart}). All leaves at or below a node with a height less than a threshold are grouped into a cluster, where the threshold is based on a typical human shoulder width, e.g. 0.6~meters.  

\item {\bf Grouping of $\depthpointsetprime$ and Cleanup.} All available observation points of  $\depthpointsetprime$ are assigned to a cluster, given that they are sufficiently close to the cluster center. Otherwise they are removed. Small clusters which originate from noise or people on the border of the field of view are removed.

\item {\bf Identifying a Cluster Representative.} For every cluster $\clusteridprime$, the point $\ped$ representing the pedestrian location of a trajectory \eqref{eq:trajectory} is selected as the point with the 95th percentile of the height $\pworldz$ in  $\clusteridprime$, defined as the person's height. 

\end{enumerate}

\begin{figure}[t]
\centering
\includegraphics[width=14cm]{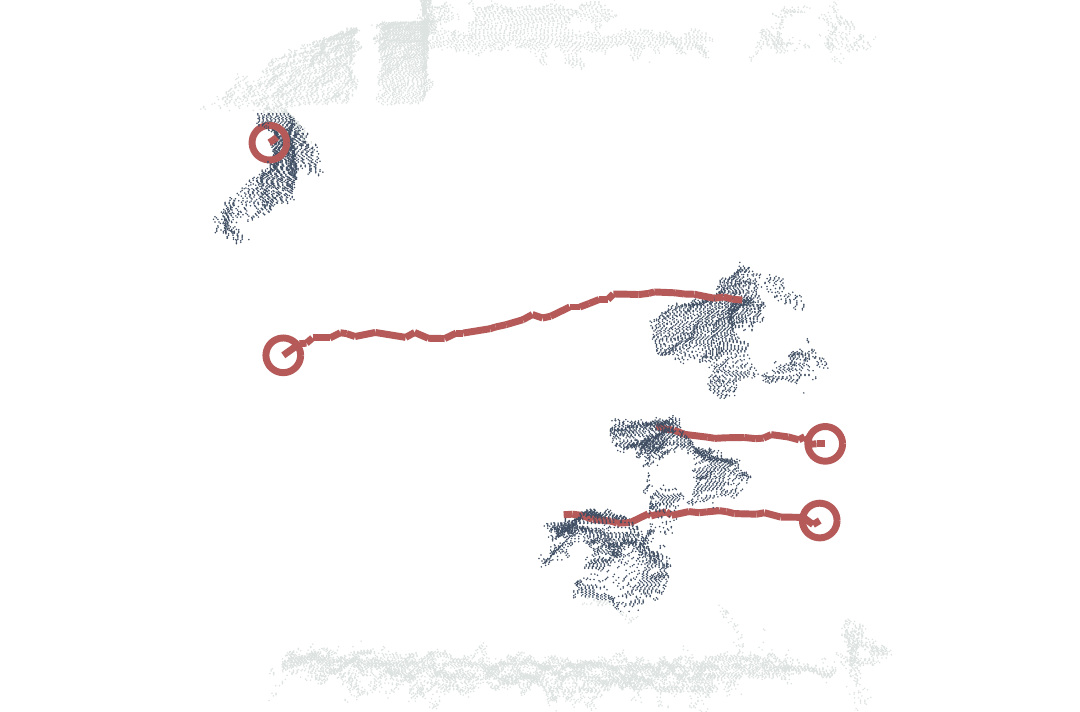}
\caption{Kinect depth raw data in 3D (walls are light gray and detected objects are dark gray) with automatic obtained trajectories (red).}
\label{fig:Tracking}
\end{figure}

This process provides robust people detections of all individuals in a single depth image. In order to obtain correspondences of multiple people over consecutive frames and hence trajectories $\trajectory$ as defined in \eqref{eq:trajectory}, a simple nearest neighbor matching is used in conjunction with a linear extrapolation from preceding frames. While other applications use more complex approaches for object tracking (see \cite{Berclaz2011} for an overview), we take advantage of the high rate of 30 frames per second provided by the Kinect. Our linear extrapolation predicts the position of individuals using the previous $n$ frames, where we chose  $n = 5$. Having the predicted location, we search for the nearest individual within a certain spatial and temporal threshold.
Figure~\ref{fig:Tracking} shows the tracking results of a short sequence.

\subsection{Tracking over Multiple Sensor Views}
\label{multipleTracking}
Combining trajectories from multiple Kinect sensors enables the observation of pedestrian movement on a larger spatial scale, which yields a richer data set for crowd modeling. Such a combination, or stitching of an individual's trajectories over multiple sensors 
requires a correct association of trajectory data from the different sensors $\kinect$. We apply a trajectory stitching approach
inspired by \cite{Stauffer2003} and \cite{Kaucic2005}.

We denote $\pedone = [t_1 \; \pedonex \; \pedoney \; \pedonez]^T$ as the first point of a pedestrian trajectory $\trajectory_i$, and 
$\pedend = [t'_N \; \pedendx \; \pedendy \; \pedendz]^T$ the last point of a pedestrian trajectory $\trajectory_j$, where $\pedonez$ and $\pedendz$ are the pedestrian height information averaged over the respective trajectory. The Euclidean distance    
\begin{equation}
d_{ij} = d(\trajectory_i, \trajectory_j) = || \pedone - \pedend || 
\label{eq:stitchingdistance}
\end{equation}
then gives an expression of the dissimilarity between trajectory end points, i.e. how unlikely it is that $\trajectory_i$ and $\trajectory_j$ were generated from the same person. Time information is expressed in seconds, and the world coordinates are denoted in meters. The four-dimensional features are therefore already in the same scale and need not be normalized. Given two trajectory sets of two Kinect sensors, the distance \eqref{eq:stitchingdistance} is used to build  a square distance matrix 
\begin{equation}
\mathbf{D} = \begin{bmatrix}
 d_{11} & d_{12} & \cdots  & d_{1n} & d_{10}^0 & \cdots  & d_{1m}^0\\ 
 d_{21} & d_{22} & \cdots  & d_{2n} & d_{20}^0 & \cdots  & d_{1m}^0\\ 
 \vdots & \vdots  &  & \vdots  & \vdots &  & \vdots \\ 
 d_{m1} & d_{m2} & \cdots  & d_{mn} & d_{m0}^0 & \cdots  & d_{mm}^0
\end{bmatrix}.
\label{eq:transitionMatrix}
\end{equation}
Here, $d_{ij}^0 > \max_{(i,j)} (d_{ij})$ denotes the null match, used when $m > n$, i.e. the number of trajectories to match between two Kinect sensors is different. 

Pairwise stitching of start and end points of trajectories to combine them to a longer trajectory can be expressed as a bipartite graph matching problem, which can be solved by the Hungarian algorithm (see \cite{Munkres1957}).
We consider distances from $\mathbf{D}$ as weights of a complete weighted bipartite graph.
From all possible trajectory matchings $C$ the Hungarian algorithm solves this assignment problem by finding the optimal assignment $C_\textup{opt}$ given by
\begin{equation}
C_\textup{opt} = \min_{C} \sum_{(i,j)\in C} d_{ij} .
\label{eq:optimalAssignment}
\end{equation}
The global optimization of the Hungarian algorithm associates all trajectory pairs, even those with very high $d_{ij}$. However, large distances $d_{ij}$ are very likely the result of interrupted or short trajectories caused by detection or tracking errors. 
We therefore apply an iterative approach as follows:

\begin{enumerate}
\item  Take into account only trajectories for which elements in the association matrix $\mathbf{D}$ in \eqref{eq:transitionMatrix} are lower than threshold $h$. This provides trajectory assignments with a very high likelihood of being correct.
\item Remove already assigned trajectories from $\mathbf{D}$, increase $h$ and calculate the assignment with the remaining trajectories. 
\item Repeat step 2 until $h$ has reached an upper boundary. Trajectories which are left without assignment cannot be matched. 
\end{enumerate}

After the trajectory matching from all sensor views, we perform a resampling and smoothing on the combined trajectories based on a cubic spline approximation. Having the Kinect sensors in a slightly overlapping setting provides a more robust similarity measure in terms of spatio-temporal relationship.

%% file: evaluation.tex
\section{Tracking Evaluation}
\label{evaluation}
Real data for pedestrian simulation calibration is often confined to trajectories which have been manually extracted from video data sets. The reason is that the required accuracy of the trajectories is very high, and often only manually extracted trajectories can fulfill such accuracy requirements. 
It is thus necessary to compare the output of the Kinect pedestrian tracking described above with the "gold standard`` of manually generated trajectories in order to have an idea how suitable automatic collection of really large data sets are. 

\subsection{Performance Evaluation of Single Sensor People Tracking }
\label{evaluationTracking}
\begin{figure}[h]
\centering
\includegraphics[width=10cm]{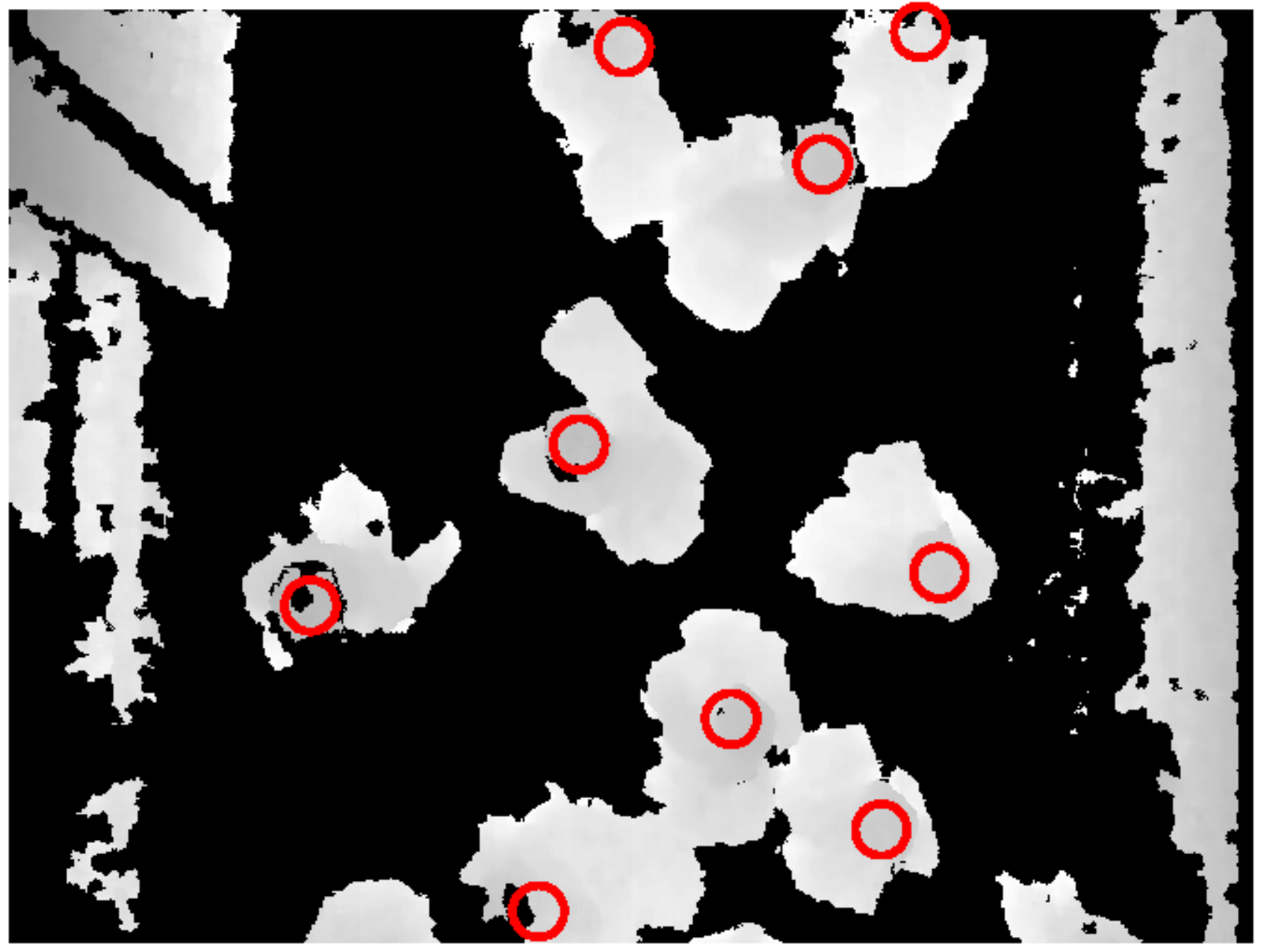}
\caption{Kinect depth raw data in 3D (gray) with manually annotated head positions of individuals (red circles).}
\label{fig:AnnotateFrame}
\end{figure}

A human observer annotated the locations of all individuals in single frames using the raw depth sensor data from the Kinect. While the Kinect's depth data does not allow for identifying persons, the body shape of individuals is still recognizable.
Our evaluation data is composed of two trajectory sets: the first data set comprises 15578 frames with pedestrian flows of low to medium density, i.e. up to 0.5 $\mbox{persons}/m^2$, and a total number of 128 persons. The second sequence includes 251 frames with a total number of 21 persons and comparably higher densities of up to 1 $\mbox{person}/m^2$. Figure \ref{fig:AnnotateFrame} illustrates a single frame from the second dataset.

\begin{figure}[htb]
  \centering
  \begin{minipage}[c]{0.49\textwidth}
    \centering
    \includegraphics[scale=0.53]{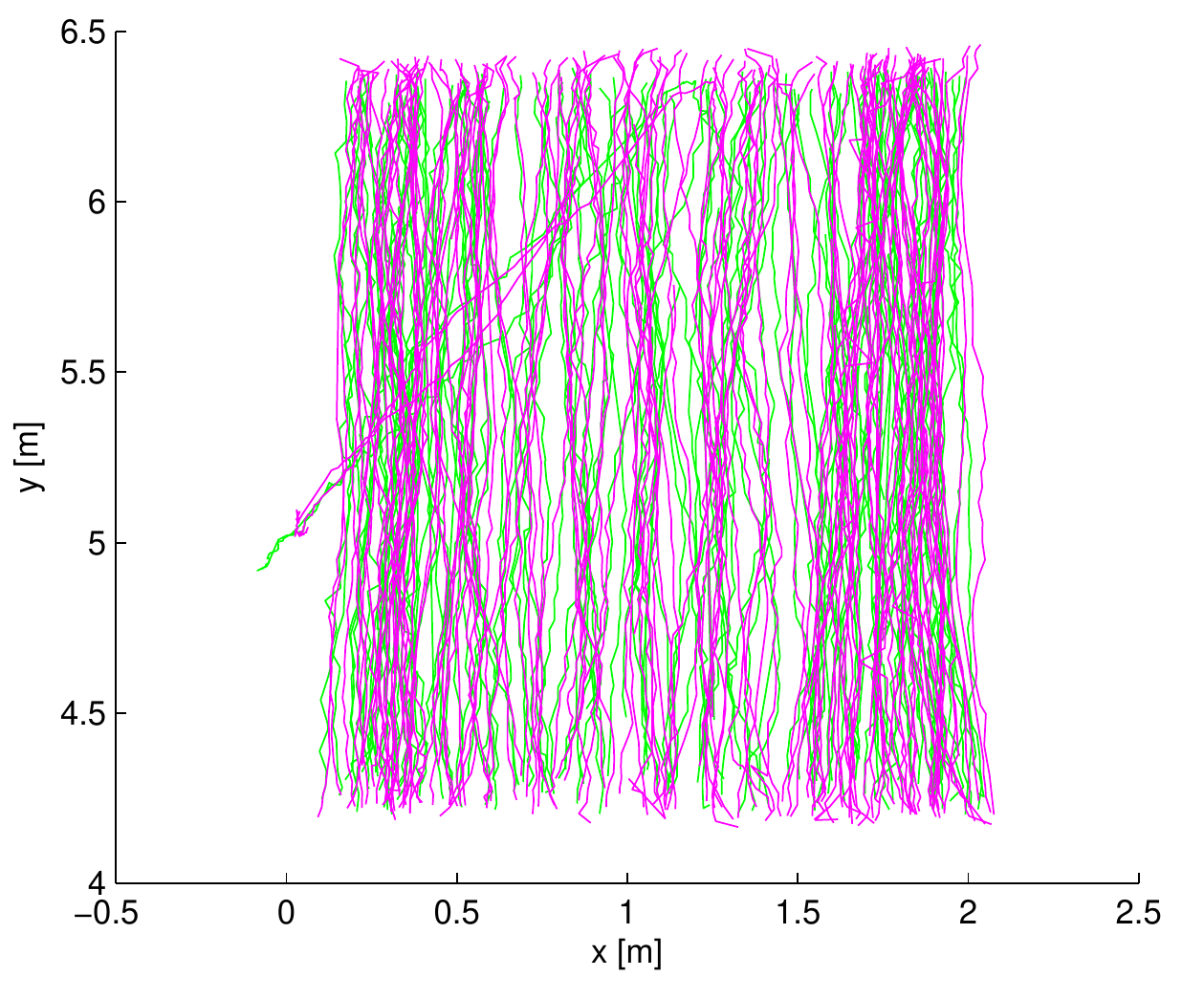}
  \\(a)\\
  \end{minipage}
  \begin{minipage}[c]{0.49\textwidth}
   \centering
    \includegraphics[scale=0.53]{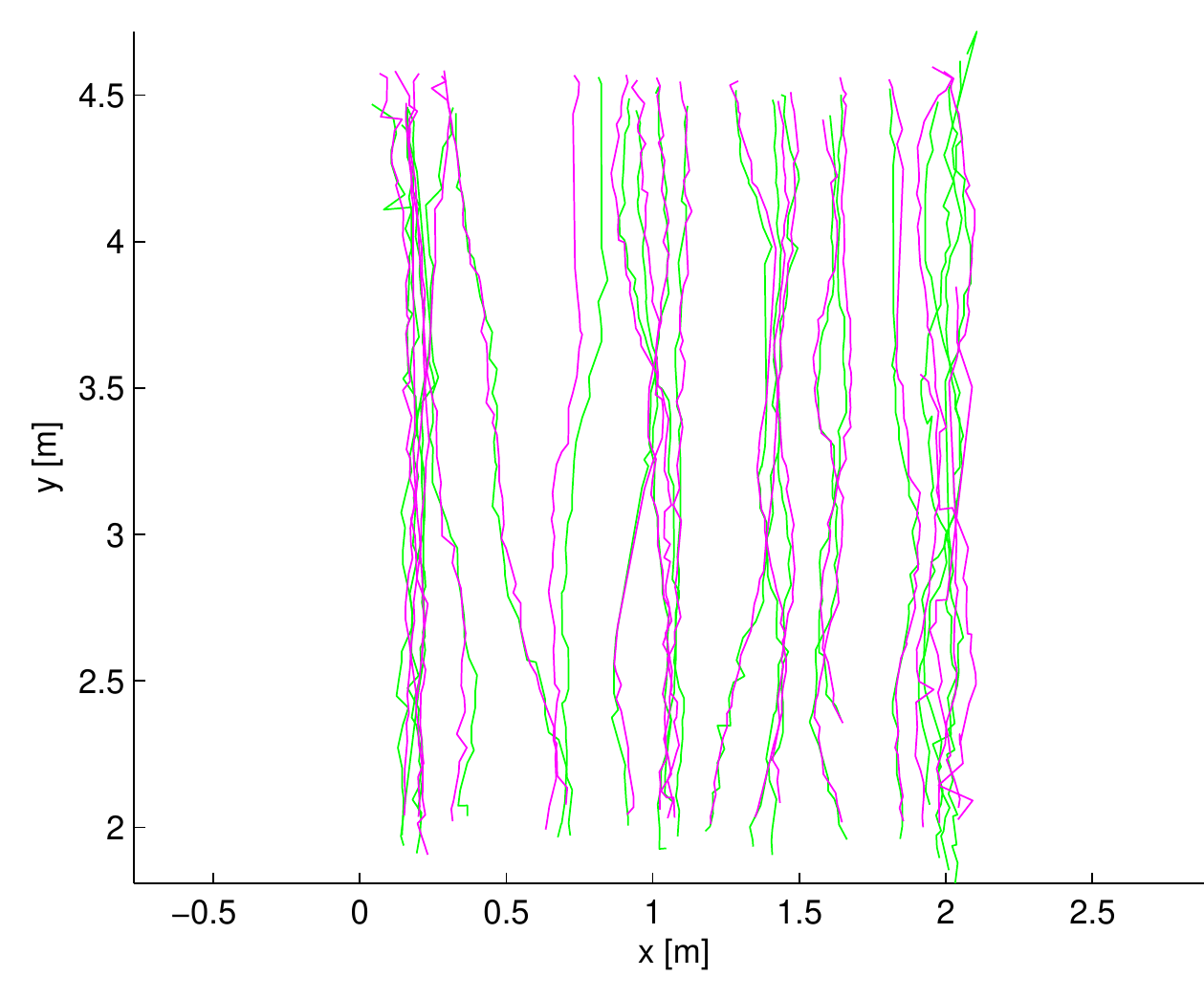}
   \\(b)\\
  \end{minipage}
  \caption{Tracking performance evaluation using ground truth (green) and automatic trajectories (magenta) including (a) 128~persons with up to 0.5 $\mbox{persons}/m^2$ and (b) 21~persons with up to 1 $\mbox{person}/m^2$. 
}
  \label{fig:TrackingPerformance}
\end{figure}

In a first step, every automatically computed trajectory $\trajectory$ is assigned to a ground truth trajectory $\trajectoryGT$ by minimizing a trajectory distance metric. Quantifying the pairwise trajectory dissimilarity in a distance metric is not trivial due to the usually different number of points. Here we used the discrete Fr\'{e}chet distance (\cite{Wien94computingdiscrete}). Following an informal interpretation, the Fr\'{e}chet distance between two trajectories is the minimum length of a leash that allows a dog and its owner to walk along their respective trajectories, from one end to the other, without backtracking. Taking into account the location and ordering of points along the trajectories,  the Fr\'{e}chet distance is well-suited for the comparison of trajectories and is less sensitive to outlier points than alternatives for arbitrary point sets such as the Hausdorff distance.

As a result of the trajectory assignment we derive a set of $P$ matching trajectory pairs for a time stamp $t$. Any remaining automatically computed trajectories which could not be matched are considered as false positives. Similarly, any remaining ground truth trajectories which could 
not
be matched are considered as misses.
Figure~\ref{fig:TrackingPerformance} shows the results based on trajectories from both sequences.
Our dataset produced zero false positives and one miss. It was seen in the data that this missed person was smaller than the defined cutoff value of 1.5 meters.
In order to quantify the position error for all correctly tracked objects over all frames, we use the Multiple Object Tracking Precision (MOTP) as described in \cite{Bernardin2008},  which is defined as
\begin{equation}
Q_\textup{MOTP} = \frac{\sum_{i,t}d_t^i}{\sum_{t}c_t},
\label{eq:MOTP}
\end{equation}
where $c_t$ is the number of matches found for time $t$. For each of these matches, $d_t^i$ denotes the discrete Fr\'{e}chet distance between the automatic and the ground truth trajectory.

The Pedestrian Detection Rate (PDR) measures the rate at which tracked pedestrians are matched to the ground truth. The value of PDR varies between 0 and 1. While 0 means poor pedestrian detection, 1 means that all ground truth pedestrians are matched. The metric is given by
\begin{equation}
Q_\textup{PDR} = \frac{\textup{TP}}{\textup{TP} + \textup{FN}},
\label{eq:PDR}
\end{equation}
where the number of matched ground truth pedestrians is denoted by true positives $\textup{TP}$. False negatives $\textup{FN}$ state the number of missing detections. Table~\ref{tb:TrackingQuality} provides the evaluation results for our detection and tracking approach.
Based on the PDR, our approach performs well on both sequences, with detection rates above 94\%. Also the localization errors stated by the MOTP are quite low.
\begin{table}[h]
  \centering
  \begin{tabular}{  l | c | c |}
    \cline{2-3}
    & $Q_\textup{PDR}$ & $Q_\textup{MOTP}$ \TS\BS \\ \hline
    \multicolumn{1}{|c|}{Sequence 1} & 96.20\% & 41.3 mm \TS\BS \\ \hline
    \multicolumn{1}{|c|}{Sequence 2} & 93.86\% & 34.0 mm \TS\BS \\ \hline
  \end{tabular}
\caption{Tracking evaluation results, showing Pedestrian Detection Rate (PDR) and Multi Object Tracking Precision (MOTP).}
\label{tb:TrackingQuality}
\end{table}

\subsection{Trajectory Stitching Performance}
\label{evaluationStitching}
\begin{figure}[t]
  \centering
  \includegraphics[scale=0.7]{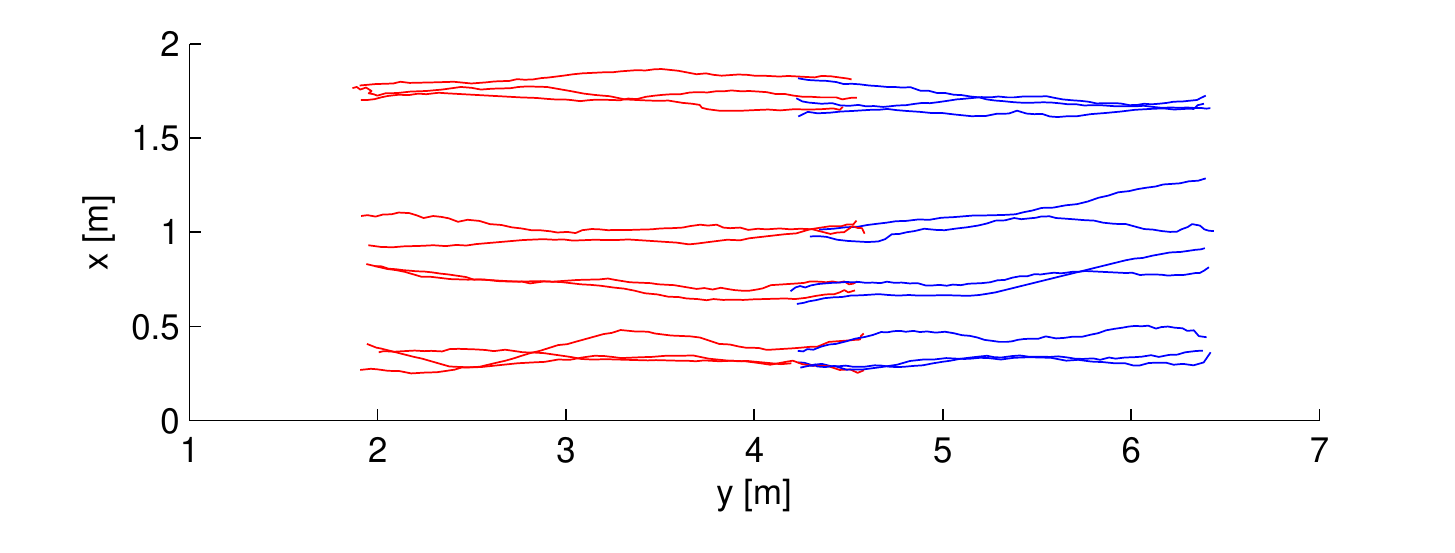}
  \caption{Association of trajectories from Kinect sensor $S_1$ (red) and $S_2$ (blue).}
  \label{fig:StitchTrajectories}
\end{figure}

The performance evaluation of our method for combining trajectories from multiple Kinect sensors is based on ground truth data with manually associated trajectories. We randomly selected two sets of automatically obtained trajectories originating from two Kinect sensors $S_1$ and $S_2$: the first data set includes 453 trajectories from $S_1$ and 442 trajectories from $S_2$ respectively. Here, a subset of 119 trajectories was manually assigned serving as ground truth data. The second data set comprises 1402 trajectories from $S_1$ and 1423 trajectories from  $S_2$ with a manually assigned subset of 50 trajectories. A selection of ten trajectories from Kinect sensors $S_1$ and $S_2$ is shown in Figure~\ref{fig:StitchTrajectories}. In this sensor setting, the trajectories are slightly overlapping which allows to derive a more robust similarity measure.

As described in Section~\ref{multipleTracking}, trajectories from both data sets were automatically combined by applying the Hungarian algorithm in an iterative manner. For both subsets we then compared the assigned trajectories derived by the automatic approach with the manual annotation. It turns out that increasing the threshold $h$ reduces the assignment quality which is documented by the True Positive Ratio (TPR) in Table~\ref{tb:StitchingQuality}. This confirms our assumption that erroneous trajectories decrease the assignment quality when applying a global optimization with the Hungarian algorithm. However, the results can be significantly improved by restricting the assignment to trajectories within a lower threshold $h$ only. Iteratively increasing $h$ enables to combine even severe interrupted or short trajectories.

\begin{table}[h]
  \centering
  \begin{tabular}{ | c | c | c | c |}
    \hline
   Threshold & Subset 1 & Subset 2 \\
$h$ & TPR & TPR \\ \hline
3	&	98.00\%	&	    99.16\%	\\	\hline
6	&	98.00\%	&	    99.16\%	\\	\hline
9	&	90.00\%	&	    99.16\%	\\	\hline
12	&	76.00\%	&	    89.08\%	\\	\hline
15	&	80.00\%	&	    91.60\%	\\	\hline
18	&	82.00\%	&	    89.92\%	\\	\hline
21	&	84.00\%	&	    89.92\%	\\	\hline
23	&	70.00\%	&	    86.55\%	\\	\hline
  \end{tabular}
\caption{Evaluation results for stitching performance.}
\label{tb:StitchingQuality}
\end{table}

%% file: results.tex
\section{Calibration of Crowd Models}
\label{results}
Crowd behavior models are used to simulate and predict how humans move around in different environments such as buildings or public spaces. In order to reflect realistic behavior, crowd behavior models must rely on empirical observations which ideally include a broad variety of human walking behavior. We performed a variety of walking experiments at the MIT's Infinite Corridor described in Section~\ref{tracking} while capturing depth image sequences of three Kinect sensors.
Applying the people tracking algorithm of Section~\ref{tracking} on the collected Kinect data sets left us with a comprehensive amount of robust trajectories. These trajectories provide the necessary information for calibrating different types of microscopic pedestrian simulation models. In the following we present experimental results of comparing three variations of the \emph{Social Force Model} (see \cite{Helbing1995}) based on our data collected for calibrating these models. 

\subsection{Walking Experiments}
\label{walkingExperiment}
We performed the walking experiments under real world conditions, meaning that the individuals crossing MIT's Infinite Corridor had no information about being observed. The main task was to calibrate different microscopic pedestrian simulation models on the {\em operational level} with relatively simple scenarios, which allow to neglect the tactical level such as route choice.  

\begin{figure}[t]
  \centering
  \begin{minipage}[c]{0.45\textwidth}
    \centering
    \includegraphics[scale=0.7]{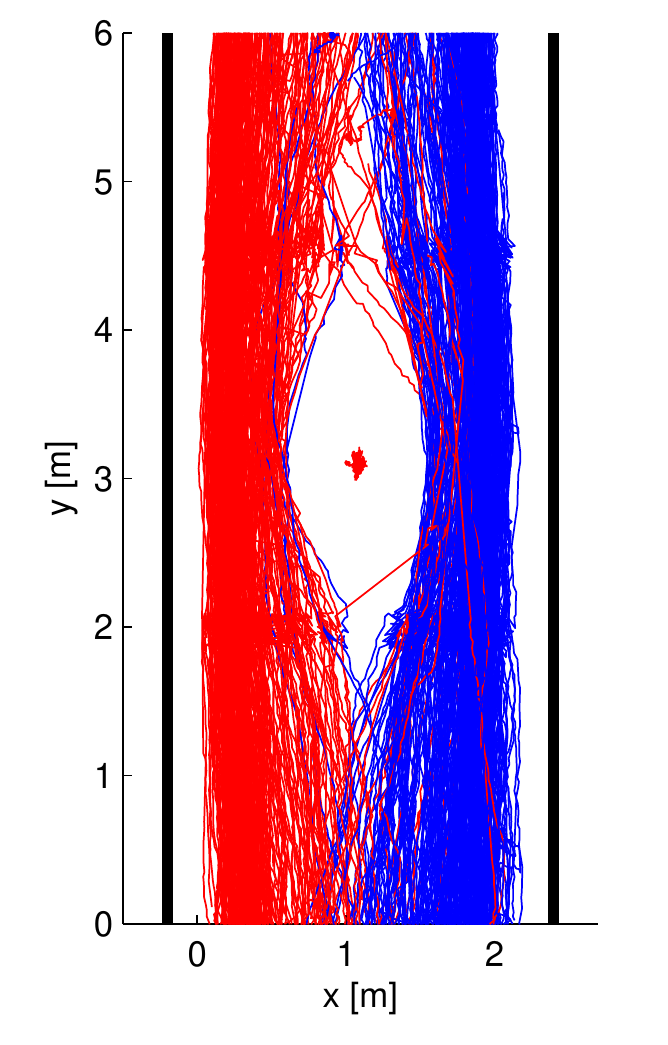}
  \\(a)\\
  \end{minipage}
  \hspace{0.5cm}
  \begin{minipage}[c]{0.45\textwidth}
   \centering
    \includegraphics[scale=0.7]{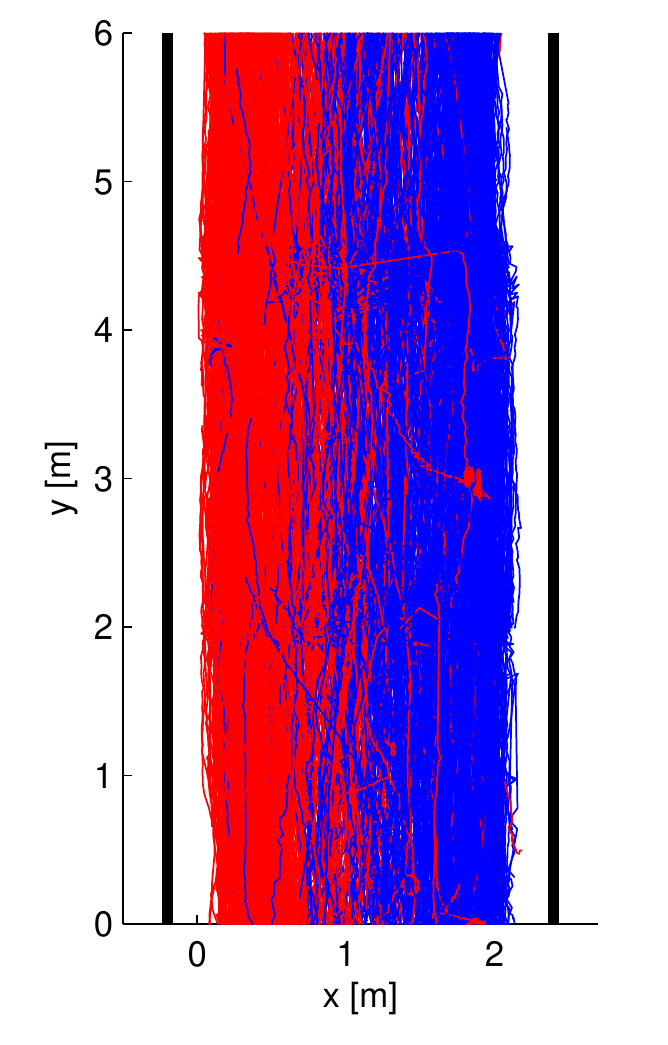}
   \\(b)\\
  \end{minipage}
  \caption{Trajectories for the calibration of crowd models automatically retrieved from (a) experiment 1 and (b) experiment 2 (walking directions are encoded in red and blue).}
  \label{fig:TrajectoriesExp}
\end{figure}

In the first walking experiment, a person standing in the center of the observed area served as an obstacle for passing people. The 558 trajectories of this setting were recorded during a period of approximately 28 minutes (see Figure~\ref{fig:TrajectoriesExp}a). 
The second walking experiment includes ``normal''  
walking behavior without any external influence for a time span of around one hour.
The 1682 trajectories computed with our Kinect approach are illustrated in Figure~\ref{fig:TrajectoriesExp}b. 
The red and blue trajectories in Figure~\ref{fig:TrajectoriesExp}a and b represent the two walking lanes in opposite directions which people form most of the time. 

\begin{figure}[t]
  \centering
  \begin{minipage}[c]{0.47\textwidth}
    \centering
    \includegraphics[scale=0.52]{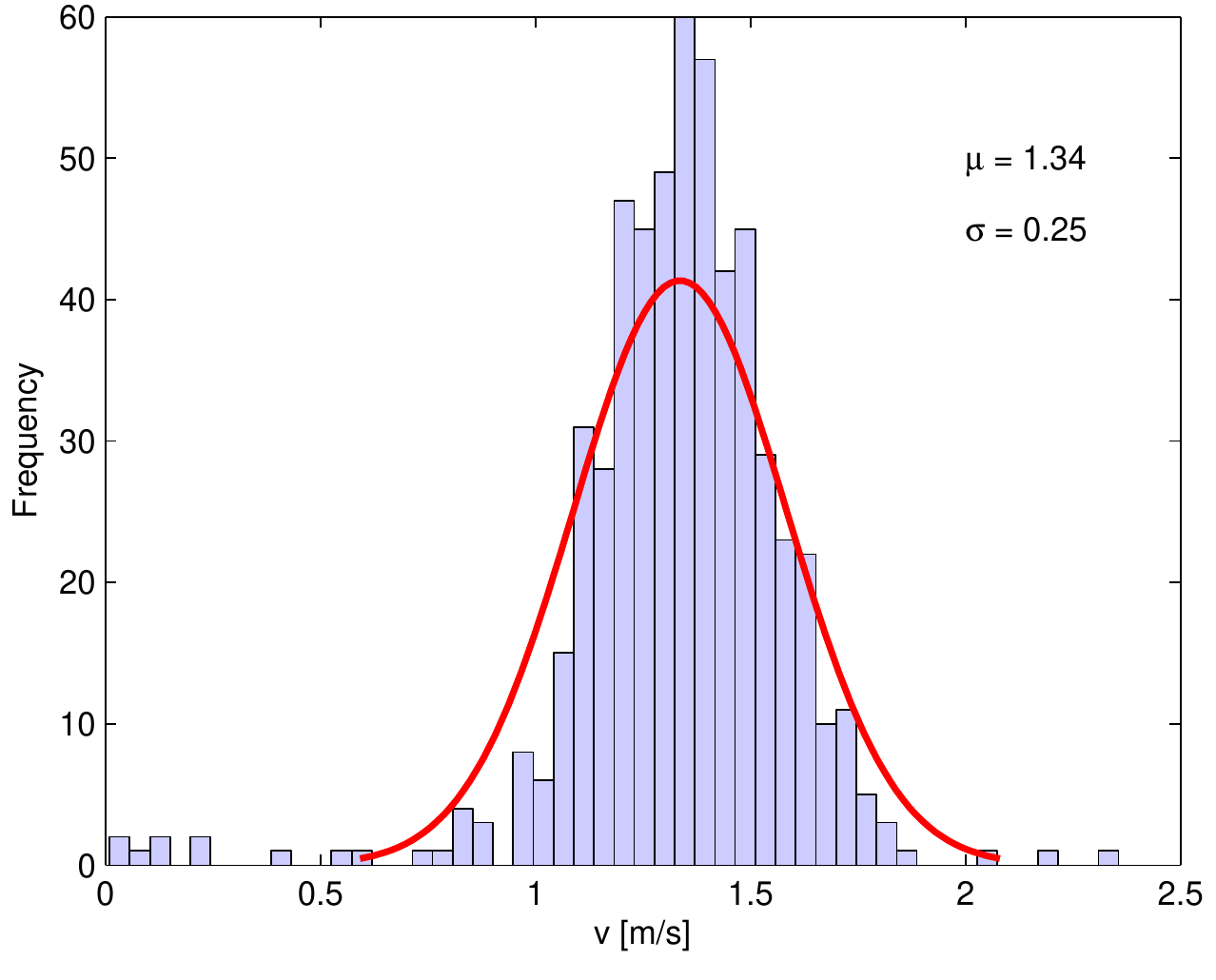}
  \\(a)\\
  \end{minipage}
  \hspace{0.05cm}
  \begin{minipage}[c]{0.47\textwidth}
   \centering
    \includegraphics[scale=0.52]{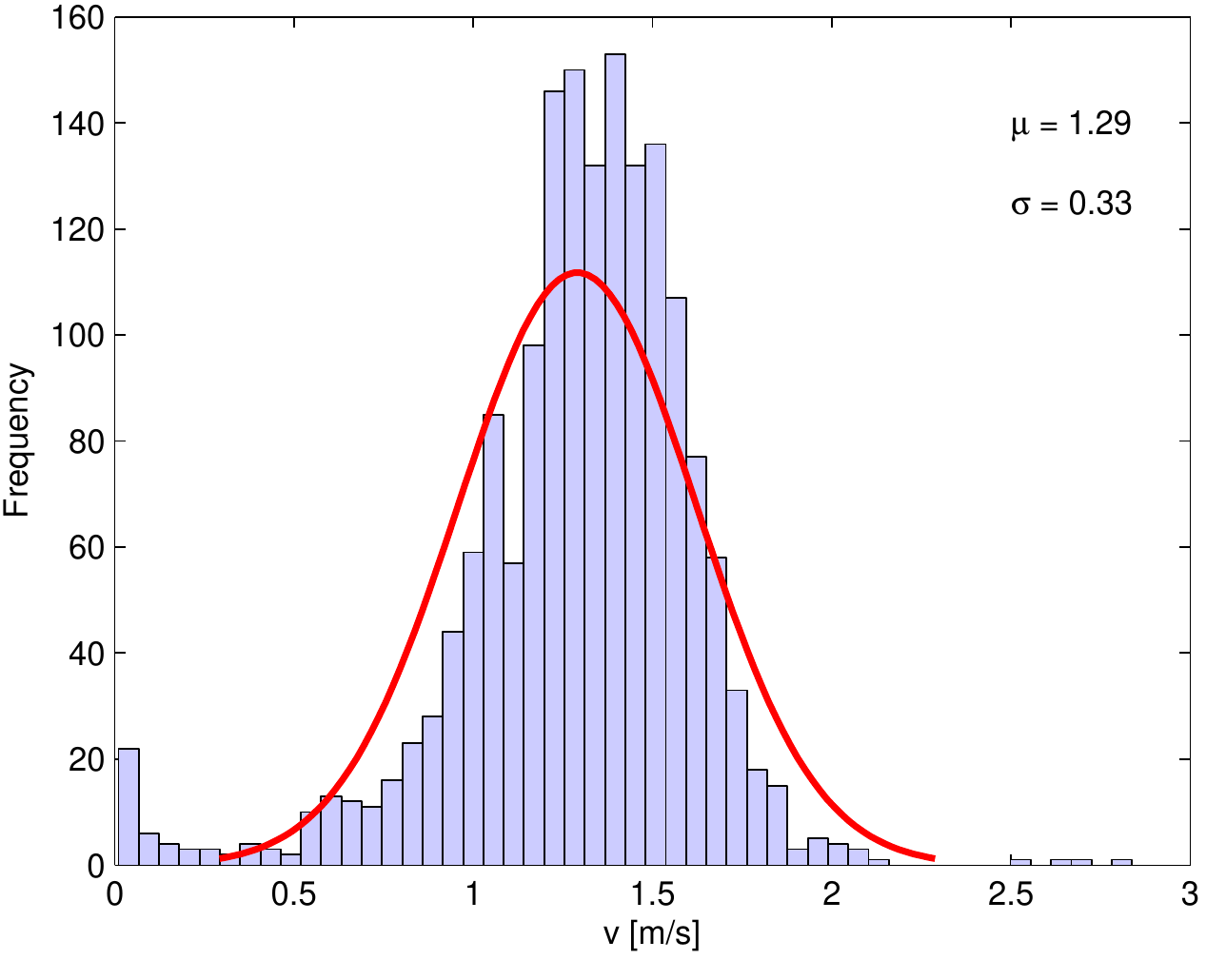}
   \\(b)\\
  \end{minipage}
  \caption{Walking speed distribution from (a) experiment 1 and (b) experiment 2.}
  \label{fig:SpeedDistribution}
\end{figure}

Figures~\ref{fig:SpeedDistribution}a and b show the walking speed histograms computed from the trajectories of the two calibration data sets (the velocity of the person acting as an obstacle in experiment 1 is filtered out). Fitted parameters of a Gaussian function to the data set result in a mean speed of 1.34 m/s and a standard deviation of 0.25 m/s. Experiment 2 shows similar results for the walking speed distribution with a mean speed of 1.29 m/s and a standard deviation of 0.33 m/s. 

\subsection{Pedestrian Simulation Model Description}
\label{modelDescription}
The models for the simulations in this work are all based on the Social Force model as presented in \cite{Helbing1995}. 
Given that movement depends on velocity and hence on acceleration, the principle of the Social Force model aims at representing individual walking behavior as a sum of different accelerations as
\begin{equation}
\forcealpha (t) = \frac{\desiredvelocity \currentdirection - \currentvelocity}{\relaxation} + \sum_{\beta\neq \alpha} \repulsiveped (t) + \sum_{i} \repulsivewall (t).
\label{eq:socialforce}
\end{equation}
The acceleration $\forcealpha$ at time $t$ of an individual $\alpha$ towards a certain goal is defined by the desired direction of movement $\currentdirection$ with a desired speed $\desiredvelocity$. 
Here, the current velocity $\currentvelocity$ is adapted to the desired speed $\desiredvelocity$ within a certain relaxation time $\relaxation$. The movement of a pedestrian $\alpha$ is influenced by other pedestrians $\beta$ which is modeled as a repulsive acceleration $\repulsiveped$. A similar repulsive behavior for static obstacles $i$ (e.g. walls) is represented by the acceleration $\repulsivewall$. For notational simplicity, we omit the dependence on time $t$ for the rest of the paper.

There exist several different formulations of the Social Force model in the literature.
We compare three variations of the Social Force model based on the general formulation~\eqref{eq:socialforce}. 

{\bf Model A}: The first model from \cite{Helbing1995} is based on a circular specification of the repulsive force given as 
\begin{equation}
\repulsiveped^A = \repulsivestrength \euler^{-\frac{\left ( r_\alpha + r_\beta - \left \| \peddistance \right \| \right )}{\repulsiverange}} \frac{\peddistance}{\left \| \peddistance \right \|},
\label{eq:repulsive1}
\end{equation}
where $r_\alpha$ and $r_\beta$ denote the radii of pedestrians $\alpha$ and $\beta$, and $\peddistance$ is the distance vector pointing from pedestrian $\alpha$ to $\beta$. The interaction of pedestrian $\alpha$ is parameterized by the strength $\repulsivestrength$ and the range $\repulsiverange$, whereas their values need to be found in the calibration process.

{\bf Model B}: The second model uses the elliptical specification of the repulsive force as described in \cite{Helbing2010} determined by
\begin{equation}
\repulsiveped^B = \repulsivestrength \euler^{-\frac{\ellipticalhelbing}{\repulsiverange}} \frac{\peddistance}{\left \| \peddistance \right \|},
\label{eq:repulsive2}
\end{equation}
where the semi-minor axis $\ellipticalhelbing$ of the elliptic formulation is given by
\begin{equation}
\ellipticalhelbing =  \frac{1}{2} \sqrt{\left ( \left \| \peddistance \right \| + \left \| \peddistance - \left (\mathbf{v}_\beta - \mathbf{v}_\alpha \right ) \Delta t \right \| \right )^2 - \left \| (\mathbf{v}_\beta - \mathbf{v}_\alpha) \Delta t\right  \|^2}.
\label{eq:elliptic}
\end{equation}
Here, the velocity vectors $\mathbf{v}_\alpha$ and $\mathbf{v}_\beta$ of pedestrians $\alpha$ and $\beta$ are included allowing to take into account the step size of pedestrians.

{\bf Model C:} The third model is an implementation of \cite{Rudloff2011} in which the repulsive force is split into one force directed in the opposite of the walking direction, i.e. the \emph{deceleration force}, and another one perpendicular to it, i.e. the \emph{evasive force}. Here, the repulsive force is given as
\begin{equation}
\repulsiveped^C = \movementdirection \underbrace{a_n \euler^{\frac{-b_n\theta_{\alpha\beta}^2}{\vrel}-c_n \left \| \peddistance \right \|}}_{\mbox{\tiny{deceleration force}}}+ \perpendiculardirection \underbrace{a_p \euler^{\frac{-b_p|\theta_{\alpha\beta}|}{\vrel}-c_p \left \| \peddistance \right \|}}_{\mbox{\tiny{evasive force}}},
\label{eq:repulsive3}
\end{equation}
where $\movementdirection$ is the direction of movement of pedestrian $\alpha$ and $\perpendiculardirection$ its perpendicular vector directing away from pedestrian $\beta$. Furthermore, $\theta_{\alpha\beta}$ is the angle between $\movementdirection$ and $\peddistance$ and $\vrel$ denotes the relative velocity between pedestrians  $\alpha$ and $\beta$.

We denote the implementations of the three above described repulsive formulations of the Social Force model as $\repulsiveped^A$, $\repulsiveped^B$ and $\repulsiveped^C$. Note that the repulsive force from static obstacles  $\repulsivewall$ is modeled by using the same functional form as given by the repulsive force from pedestrians. Here, the point of an obstacle $i$ closest to pedestrian $\alpha$ replaces the position $\beta$ and ${\mathbf v}_i$ is set to zero. Furthermore, we take into account that pedestrians have a higher response to other pedestrians in front of them by including an anisotropic behavior, as described in \cite{Helbing2010}, into the first two formulations. 

\subsection{Model Calibration}
\label{modelCalibration}
The process of model calibration involves the finding of parameter values which produce realistic crowd behavior in the simulation results. We estimated values for the different parameters in the three described model approaches $\repulsiveped^A$, $\repulsiveped^B$ and $\repulsiveped^C$ based on our empirical data set from the walking experiments. 
The trajectory data were divided into a non-overlapping calibration and validation data set (validation is described in Section~\ref{validationResults}) as shown in Table~\ref{tb:TrajectoryDataset}.

\begin{table}[h]
  \centering
  \begin{tabular}{  l | c | c |}
    \cline{2-3}
    & \multicolumn{2}{|c|}{Number of trajectories} \TS\BS \\ \cline{2-3}
    & Experiment 1 & Experiment 2 \TS\BS \\ \hline
    \multicolumn{1}{|c|}{Calibration Set} &424 & 1121 \TS\BS \\ \hline
    \multicolumn{1}{|c|}{Validation Set} & 134 & 561 \TS\BS \\ \hline
    \multicolumn{1}{|c|}{Total} & 558 & 1682 \TS\BS \\ \hline
  \end{tabular}
\caption{Partitioning of the trajectory data set for model calibration and validation.}
\label{tb:TrajectoryDataset}
\end{table}

The literature describes different techniques for calibrating microscopic simulation models: one way is to estimate parameter values directly from the trajectory data by extracting pedestrian's acceleration (\cite{Hoogendoorn2006}). However, as shown in \cite{Rudloff2011} this method has several drawbacks, even with small errors in the trajectories. For instance, using the acceleration instead of the spatial position introduces a significant noise due to the second derivative. Furthermore, this might lead to error-in-variables problems and parameter estimates possibly result in a bias towards zero.

Our calibration uses a simulation approach, where each pedestrian is simulated separately while keeping the remaining pedestrians on their observed trajectory. Each simulation run is performed according to the following procedure: the position and the desired goal for a simulated pedestrian $\alpha$ are extracted from the start and end point of the associated observed trajectory $\trajectoryalpha$. 
The desired velocity $\desiredvelocity$ of pedestrian $\alpha$ is defined as the 90th percentile of the observed velocities. 
The magnitude of the current velocity vector $\currentvelocity$ is set equal to $\desiredvelocity$, directing towards the pedestrian's desired goal. Pedestrian $\alpha$ is simulated for $M_{\alpha} = \left | \trajectoryalpha \right |$ timesteps during  time $t$, with $t_\alpha^{in} \le t \le  t_\alpha^{out}$, where both bounds are again derived from the observed trajectory.

After having simulated a set of $N$ pedestrians from the calibration data set with the above procedure, a similarity measure $s$ for testing the fit of our simulated trajectories can be computed as
\begin{equation}
s = \frac{1}{N}\sum_{\alpha=1}^{N}\left ( \frac{d(\alpha)}{t_\alpha^{out} - t_\alpha^{in}} +g(\alpha) \right ).
\label{eq:similarityMeasure}
\end{equation}
For a pedestrian $\alpha$, the mean Euclidean distance 
\begin{equation}
d(\alpha) = d(\trajectoryalpha,\trajectoryalphaprime) = \frac{1}{M_{\alpha}}\sum_{i=1}^{M_{\alpha}} \left \| \point_\alphai - \point_{\alphai}' \right \|
\label{eq:trajectoryDistance}
\end{equation}
provides the dissimilarity between positions $\point_\alphai = [t_\alphai, x_\alphai, y_\alphai]^T$ of the observed trajectory $\trajectory_\alpha$ and positions $\point_{\alphai}' = [t_\alphai', x_\alphai', y_\alphai']^T$ of the simulated trajectory $\trajectory_{\alpha}'$. Furthermore, the length of trajectories is defined by $\left | \trajectoryalpha \right | = \left | \trajectoryalphaprime \right | = M_{\alpha}$.
Since none of the used models explicitly restricts overlapping between pedestrians, an overlap penalty is added denoted by
\begin{equation}
g(\alpha) = \frac{1}{N-1}\sum_{\beta\neq \alpha}\max_{\substack{t}} \left (0, \frac{1}{\left \| \peddistance(t) \right \|} + \frac{1}{r_\alpha + r_\beta} \right ).
\label{eq:penaltyFunction}
\end{equation}

Model parameter values are estimated by applying an optimization algorithm to find the best possible fit by minimizing the objective function \eqref{eq:similarityMeasure}. We use a genetic algorithm which does not suffer from a starting value problem to find the neighborhood of the global minimum. The estimated parameter values obtained by the genetic algorithm are then used as initial values for the Nelder-Mead algorithm (see \cite{Lagarias1998}) to refine the result. 
This hybrid approach allows finding the global minimum while being numerically efficient.

\subsection{Validation Results}
\label{validationResults}
The results for the parameter fit of the individual models are provided in Table~\ref{tb:ParameterFit} as $s_\textup{cal}$ for the calibration data set and $s_\textup{val}$ for the validation data set. 
The best possible value for~\eqref{eq:similarityMeasure} is $s = 0$. For both experiments, the best fit of the objective function with the compared modeling approaches could be achieved using the repulsive formulation from  $\repulsiveped^C$ defined in~\eqref{eq:repulsive3}. 
\begin{table}[h]
  \centering
  \begin{tabular}{ l | c | c | c | c | c | c |}
    \cline{2-7}
    & \multicolumn{3}{|c|}{Experiment 1} & \multicolumn{3}{|c|}{Experiment 2} \TS\BS \\ \cline{2-7}
    & $\repulsiveped^A$ & $\repulsiveped^B$ & $\repulsiveped^C$ & $\repulsiveped^A$ & $\repulsiveped^B$ & $\repulsiveped^C$ \TS\BS \\ \hline
    \multicolumn{1}{|l|}{$s_\textup{cal}$} & 0.0951 & 0.0887 & 0.0640 & 0.0932 & 0.0925 & 0.0820 \TS\BS \\ \hline
    \multicolumn{1}{|l|}{$s_\textup{val}$} & 0.1439 & 0.0927 & 0.0826 & 0.1017 & 0.0996 & 0.0929 \TS\BS \\ \hline
  \end{tabular}
\caption{Fit of the parameter values for three different Social Force formulations based on calibration and validation data set.}
\label{tb:ParameterFit}
\end{table}

\begin{figure}[t]
  \centering
  \begin{minipage}[c]{0.29\textwidth}
    \centering
    \includegraphics[scale=0.6]{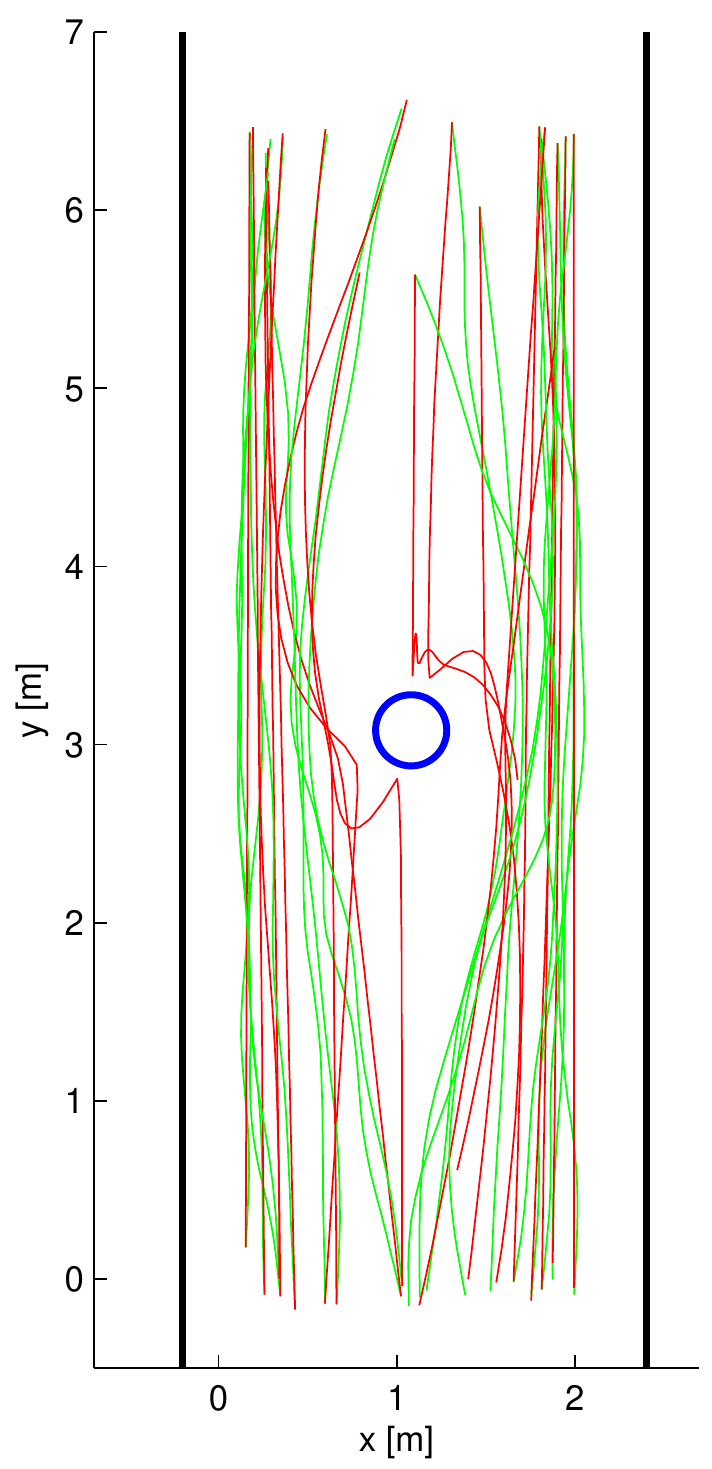}
    \\(a)\\
  \end{minipage}
  \hspace{0.3cm}
  \begin{minipage}[c]{0.29\textwidth}
   \centering
    \includegraphics[scale=0.6]{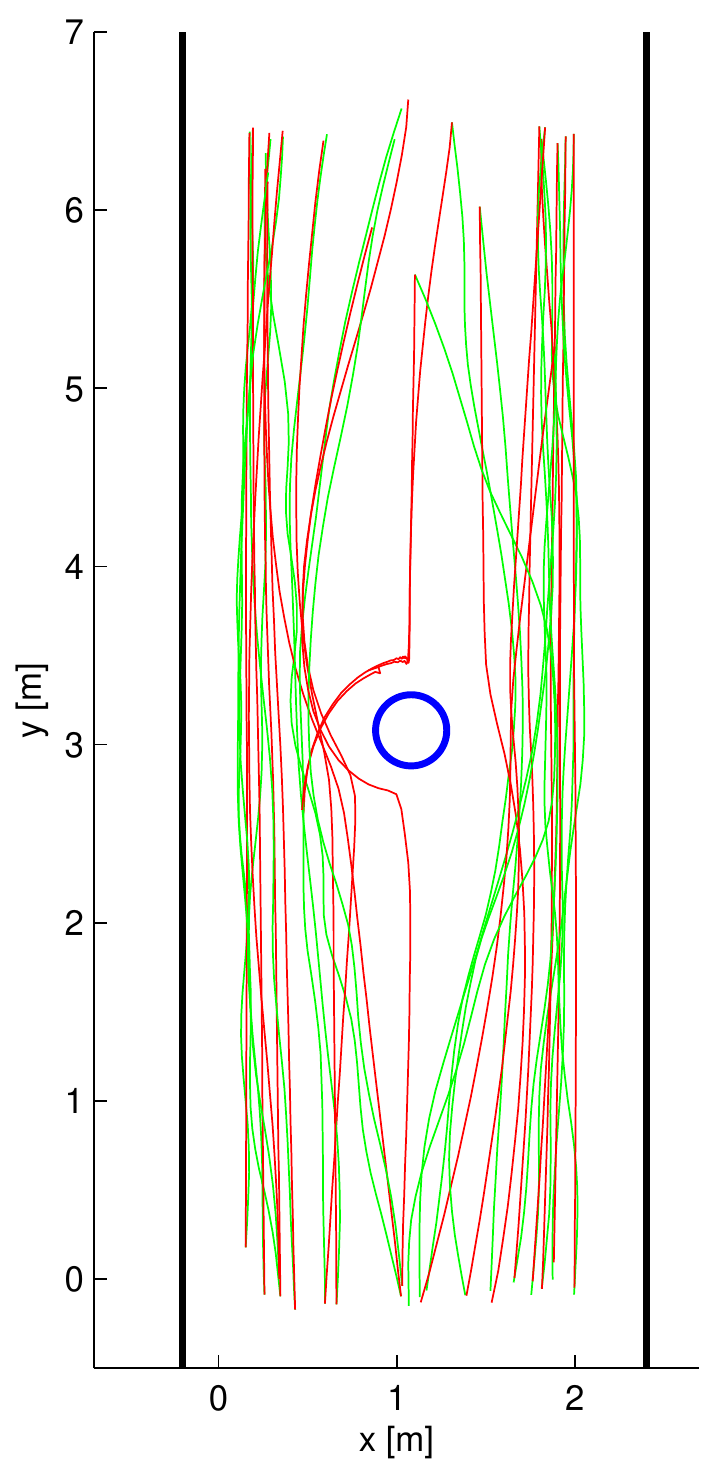}
    \\(b)\\
  \end{minipage}
  \hspace{0.3cm}
  \begin{minipage}[c]{0.29\textwidth}
   \centering
    \includegraphics[scale=0.6]{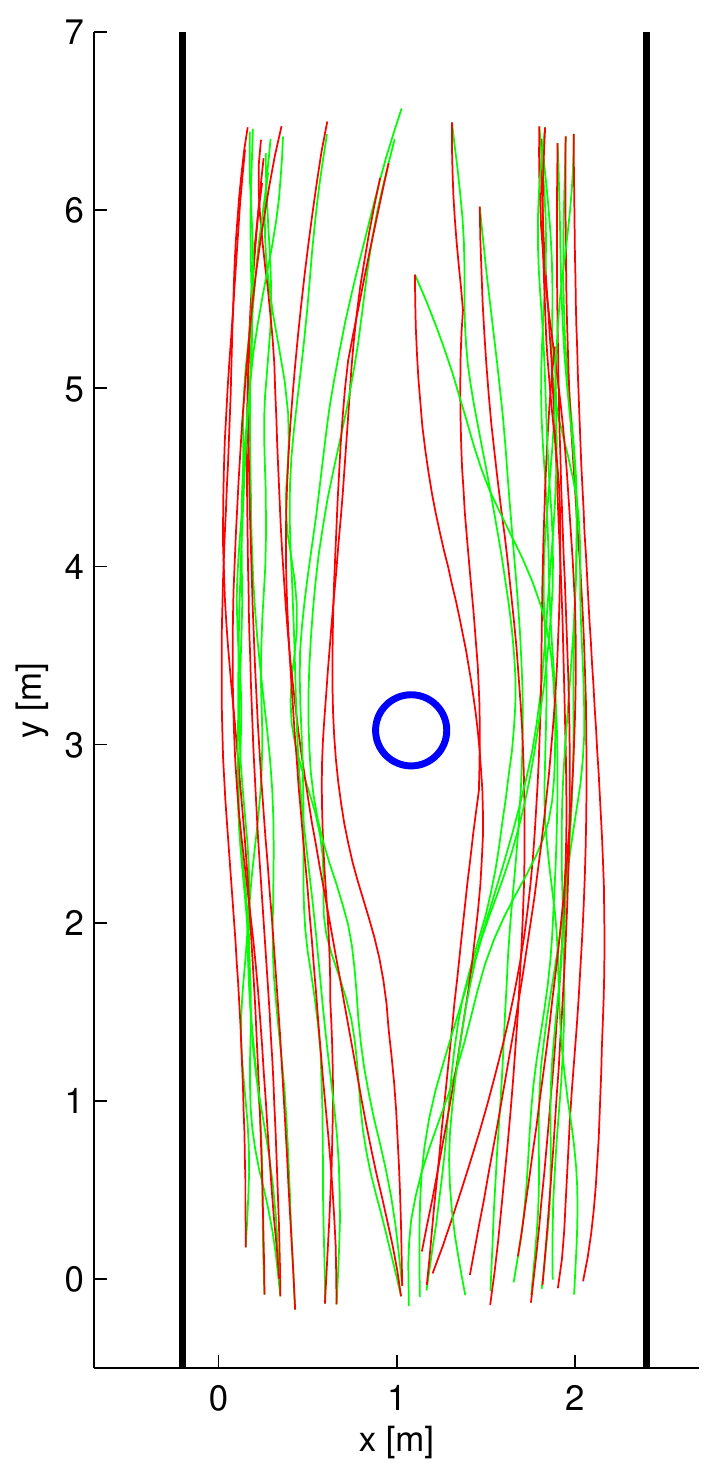}
    \\(c)\\
  \end{minipage}
  \hspace{0.2cm}
  \caption{Validation results of different Social Force models showing observed (green) and simulated trajectories (red) using (a) $\repulsiveped^A$, (b) $\repulsiveped^B$ and (c) $\repulsiveped^C$ as repulsive force.}
  \label{fig:CalibrationResultsSingleShort}
\end{figure}

By applying the three Social Force models on only a small subset of our validation data set, their basic ability of representing crowd behavior can be evaluated in a qualitative manner. Figure~\ref{fig:CalibrationResultsSingleShort} shows the results of a simulation run with 19 pedestrians in the setting of experiment~1: the simulation results of the circular force formulation from $\repulsiveped^A$ in Figure~\ref{fig:CalibrationResultsSingleShort}a indicate that simulated pedestrians evade relatively late with a strong deceleration caused by the static person in the center. To avoid running into the obstacle some pedestrians even move slightly backward from the obstacle. 
This collision avoidance behavior differs significantly from the observed trajectories. As illustrated in Figure~\ref{fig:CalibrationResultsSingleShort}b, the walking behavior from the simulations with $\repulsiveped^B$ is less abrupt as a result of the included velocity dependence. However, pedestrian deceleration is again unrealistically strong when individuals directly approach the static obstacle. From a qualitative point of view, simulation results obtained by using $\repulsiveped^C$ exhibit the best results in our comparison (see Figure~\ref{fig:CalibrationResultsSingleShort}c). Separating the forces into a deceleration and an evasive component results in individual trajectories which match very well with the observations.

\begin{figure}[t]
  \centering
  \begin{minipage}[c]{0.47\textwidth}
    \centering
    \includegraphics[scale=0.39]{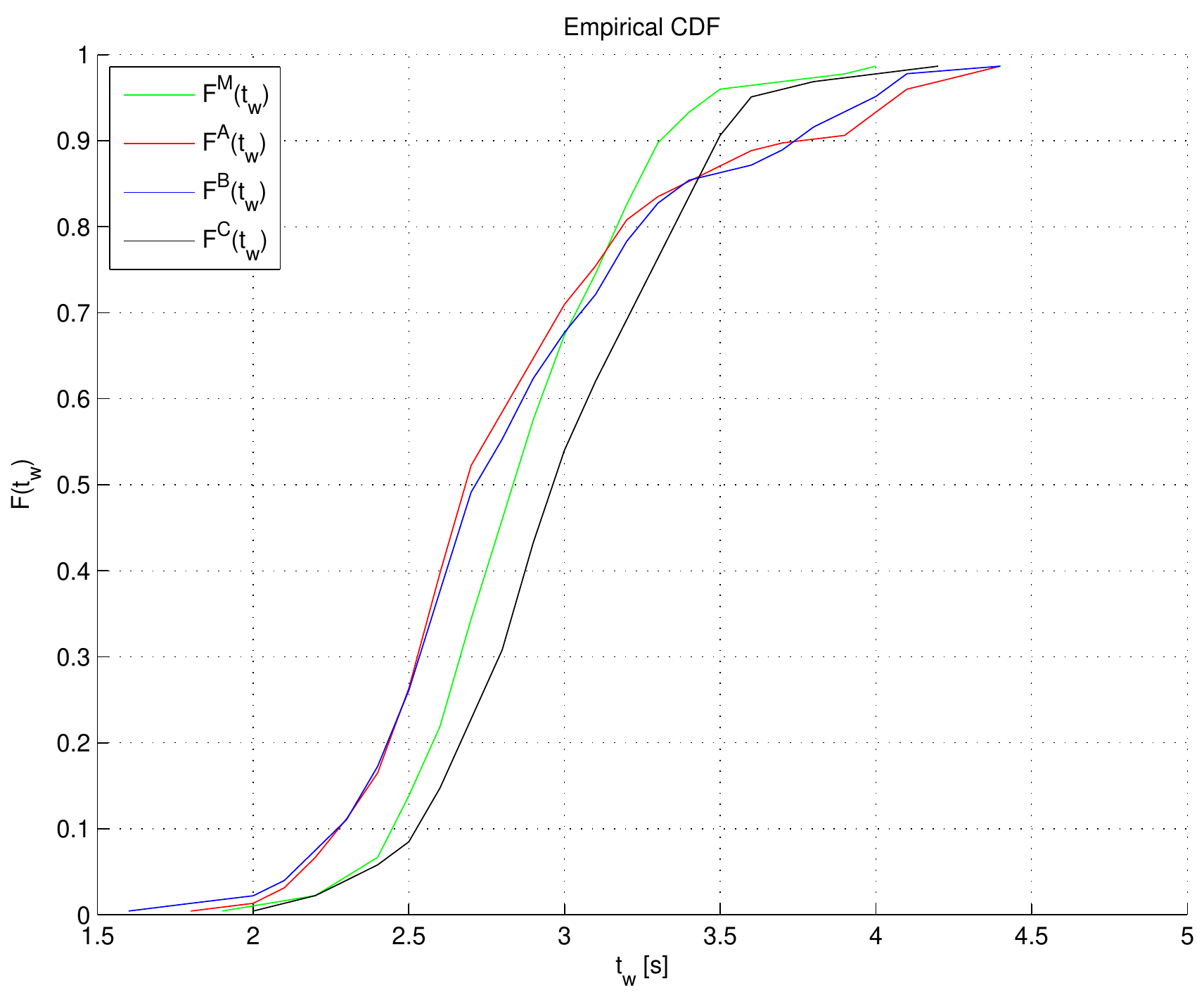}
  \\(a)\\
  \end{minipage}
  \hspace{0.1cm}
  \begin{minipage}[c]{0.47\textwidth}
   \centering
    \includegraphics[scale=0.39]{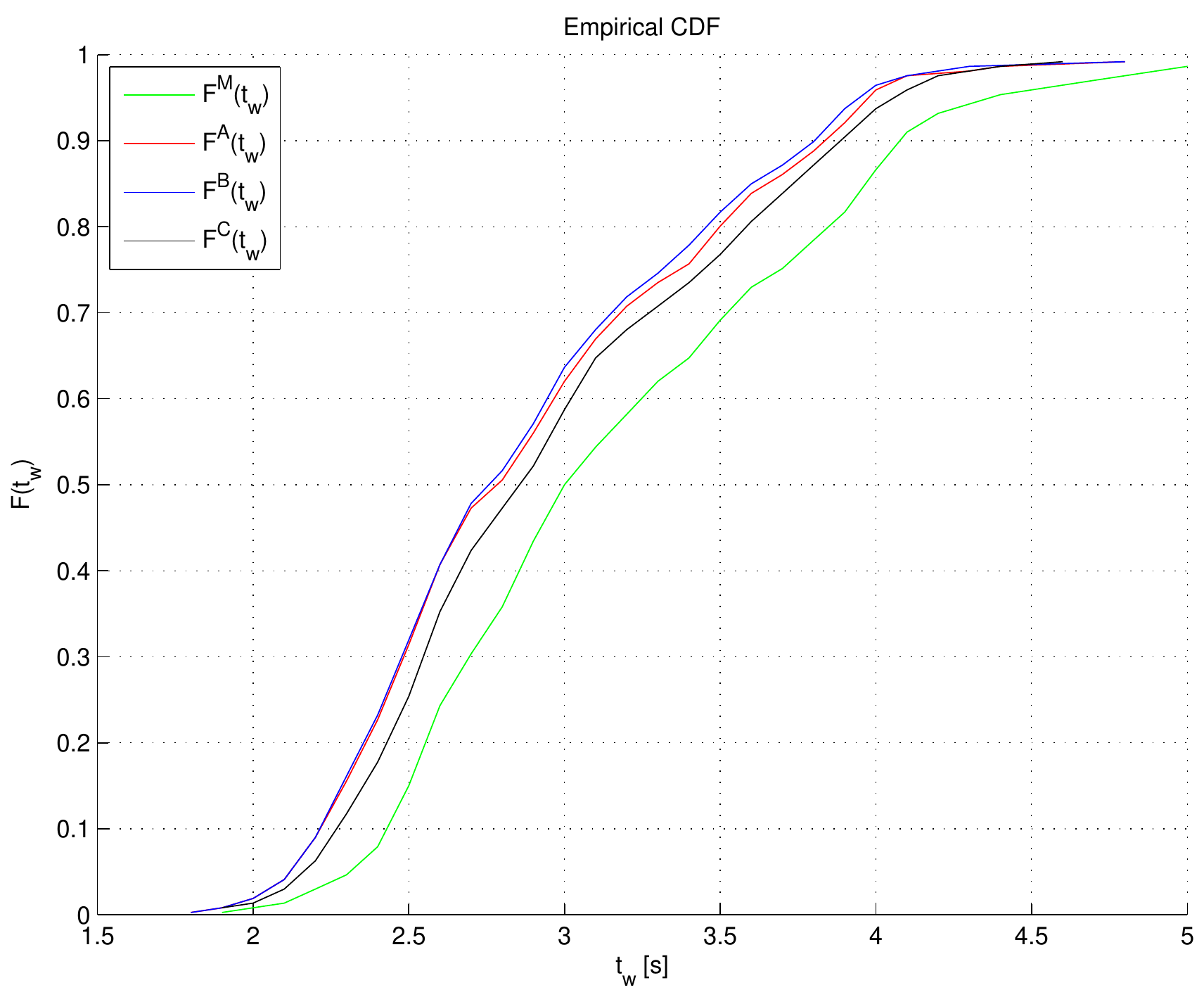}
   \\(b)\\
  \end{minipage}
  \hspace{0.1cm}
  \caption{Measured and simulated walking time distributions from (a) experiment 1 and (b) experiment 2.}
  \label{fig:WalkingTimeDistribution}
\end{figure}

For capacity estimations in infrastructures the walking times of pedestrians are of particular importance. Accordingly, pedestrian simulation models need to be able to reproduce realistic walking times even if they are not specifically calibrated for this purpose. Since the models in this work were calibrated using the similarity of trajectories as the objective function, we also want to evaluate their ability to correctly predict the walking time distribution based on our validation data set. Figure~\ref{fig:WalkingTimeDistribution} shows the cumulative distribution functions of walking times $t_w$ derived from measured $F^M$ and simulated $F^A$, $F^B$, $F^C$ trajectories provided by $\repulsiveped^A$, $\repulsiveped^B$, $\repulsiveped^C$ respectively. The results for experiment 1 (see Figure~\ref{fig:WalkingTimeDistribution}a) demonstrate that the circular $F^A$ and elliptical $F^B$ formulation for the repulsive force in the Social Force model significantly deviate from the measured walking time distribution $F^M$.  However, the formulation used to derive $F^C$ provides a good replication of the measured walking time distribution $F^M$. In order to support this finding, we used a two-sample Kolmogorov-Smirnov test (see \cite{Massey51}) to compare each walking time distribution from the simulations with the measured distribution $F^M$. For a significance level of 0.05, we can reject the null hypothesis that $F^A$ and $F^M$ as well as $F^B$ and $F^M$ are from the same continuous distribution. However, the null hypothesis holds when comparing $F^C$ and $F^M$.

%% file: conclusion.tex
\section{Conclusion}
\label{conclusion}
In this work we have developed algorithms to use the Microsoft Kinect -- basically a camera that also records 3-dimensional information in the form of a depth image -- for automatic data collection of crowd movement from an elevated view. We have shown that the use of the Kinect allows the automated capture of human motion trajectories with high accuracy, overcoming many limitations of methods that have been applied so far.
The scanning area is scalable by combining multiple Kinects, thus allowing high flexibility for measurements in different environments. We applied our tracking algorithm to collect an extensive data set in the MIT's Infinite Corridor for calibrating and comparing three variations of the Social Force model.

In order to capture human motion trajectories throughout the sensing areas of multiple Kinects, the depth information from individual Kinect sequences is mapped into a common world coordinate system using a rigid transformation. 
Our approach groups depth information from a single Kinect in the world coordinate system into individual pedestrians based on hierarchical clustering. These detections are tracked over time to obtain individual trajectories.

Evaluating the detection performance with two manually annotated ground truth data sets shows a Pedestrian Detection Rate of 94\% and 96\%, respectively. The position error for all correctly tracked objects is quantified as Multiple Object Tracking Precision and reveals relatively small values of around 4~cm. In order to observe pedestrians on a larger spatial scale, we developed methods for combining pedestrian trajectories from multiple Kinect sensors. Again, we evaluated our trajectory stitching with manually annotated ground truth data sets and received a True Positive Ratio of up to 98\%. In conclusion, our tracking approach is capable of delivering trajectories with an accuracy which we consider sufficient for calibrating microscopic pedestrian simulation models. In the future our approach could be extended in order to also estimate the orientation of body parts, i.e. head and shoulder pose. This would allow us to gain more data on how humans perceive and interact with their environment which is particularly useful for evaluating visual information systems, such as guidance systems or lights.

By applying our tracking approach in two walking experiments performed under real world conditions in the MIT's Infinite Corridor, we gathered a total of 2240 trajectories. We compared three variations of the Social Force model by calibrating them with our trajectory data. The validation results revealed that collision avoidance behavior in the Social Force model can be improved by including the relative velocity between individuals. Furthermore, dividing the repulsive force into a deceleration and an evasion part delivered the best quantitative and qualitative results out of the investigated models. However, dividing the repulsive force leads to a larger number of parameters, which makes the calibration process itself more complex and computationally expensive.
For future work we will increase our data set by obtaining trajectories under various experimental settings, such as involving different forms of obstacles.
Going forward we believe that the adoption of the Kinect could be extremely useful for the development and calibration of crowd models -- but also as a tool to better understand human crowd behavior and hence provide invaluable input to the design of all those spaces that need to respond to it -- starting from our cities.

%% file: TRC_KinectsAndHumanKinetics.bbl
\begin{thebibliography}{32}
\expandafter\ifx\csname natexlab\endcsname\relax\def\natexlab#1{#1}\fi
\expandafter\ifx\csname url\endcsname\relax
  \def\url#1{\texttt{#1}}\fi
\expandafter\ifx\csname urlprefix\endcsname\relax\def\urlprefix{URL }\fi
\providecommand{\eprint}[2][]{\url{#2}}
\providecommand{\bibinfo}[2]{#2}
\ifx\xfnm\relax \def\xfnm[#1]{\unskip,\space#1}\fi
\bibitem[{Antonini et~al.(2006)Antonini, Bierlaire and Weber}]{Antonini2006}
\bibinfo{author}{Antonini, G.}, \bibinfo{author}{Bierlaire, M.},
  \bibinfo{author}{Weber, M.}, \bibinfo{year}{2006}.
\newblock \bibinfo{title}{{D}iscrete {C}hoice {M}odels of {P}edestrian
  {W}alking {B}ehavior}.
\newblock \bibinfo{journal}{Transportation Research Part B: Methodological}
  \bibinfo{volume}{40}, \bibinfo{pages}{667 -- 687}.
\bibitem[{Berclaz et~al.(2011)Berclaz, Fleuret, Turetken and Fua}]{Berclaz2011}
\bibinfo{author}{Berclaz, J.}, \bibinfo{author}{Fleuret, F.},
  \bibinfo{author}{Turetken, E.}, \bibinfo{author}{Fua, P.},
  \bibinfo{year}{2011}.
\newblock \bibinfo{title}{{Multiple Object Tracking Using K-Shortest Paths
  Optimization}}.
\newblock \bibinfo{journal}{IEEE Transactions on Pattern Analysis and Machine
  Intelligence (PAMI)} \bibinfo{volume}{33}, \bibinfo{pages}{1806 --1819}.
\bibitem[{Bernardin and Stiefelhagen(2008)}]{Bernardin2008}
\bibinfo{author}{Bernardin, K.}, \bibinfo{author}{Stiefelhagen, R.},
  \bibinfo{year}{2008}.
\newblock \bibinfo{title}{{Evaluating Multiple Object Tracking Performance: The
  CLEAR MOT Metrics}}.
\newblock \bibinfo{journal}{EURASIP Journal on Image and Video Processing}
  \bibinfo{volume}{2008}, \bibinfo{pages}{1:1--1:10}.
\bibitem[{Berrou et~al.(2007)Berrou, Beecham, Quaglia, Kagarlis and
  Gerodimos}]{Berrou2005}
\bibinfo{author}{Berrou, J.}, \bibinfo{author}{Beecham, J.},
  \bibinfo{author}{Quaglia, P.}, \bibinfo{author}{Kagarlis, M.},
  \bibinfo{author}{Gerodimos, A.}, \bibinfo{year}{2007}.
\newblock \bibinfo{title}{{Calibration and Validation of the Legion Simulation
  Model using Empirical Data}}, in: \bibinfo{editor}{Waldau, N.},
  \bibinfo{editor}{Gattermann, P.}, \bibinfo{editor}{Knoflacher, H.},
  \bibinfo{editor}{Schreckenberg, M.} (Eds.), \bibinfo{booktitle}{Proceedings
  of the Conference on Pedestrian and Evacuation Dynamics (PED 2005)},
  \bibinfo{publisher}{Springer}, \bibinfo{address}{Berlin, Heidelberg}. pp.
  \bibinfo{pages}{167--181}.
\bibitem[{Boltes et~al.(2010)Boltes, Seyfried, Steffen and
  Schadschneider}]{Boltes2008}
\bibinfo{author}{Boltes, M.}, \bibinfo{author}{Seyfried, A.},
  \bibinfo{author}{Steffen, B.}, \bibinfo{author}{Schadschneider, A.},
  \bibinfo{year}{2010}.
\newblock \bibinfo{title}{{Automatic Extraction of Pedestrian Trajectories from
  Video Recordings}}, in: \bibinfo{editor}{Klingsch, W.W.F.},
  \bibinfo{editor}{Rogsch, C.}, \bibinfo{editor}{Schadschneider, A.},
  \bibinfo{editor}{Schreckenberg, M.} (Eds.), \bibinfo{booktitle}{Proceedings
  of the Conference on Pedestrian and Evacuation Dynamics (PED 2008)},
  \bibinfo{publisher}{Springer}, \bibinfo{address}{Berlin, Heidelberg}. pp.
  \bibinfo{pages}{43--54}.
\bibitem[{Breitenstein et~al.(2011)Breitenstein, Reichlin, Leibe, Koller-Meier
  and Van~Gool}]{Breitenstein2011}
\bibinfo{author}{Breitenstein, M.D.}, \bibinfo{author}{Reichlin, F.},
  \bibinfo{author}{Leibe, B.}, \bibinfo{author}{Koller-Meier, E.},
  \bibinfo{author}{Van~Gool, L.}, \bibinfo{year}{2011}.
\newblock \bibinfo{title}{{Online Multi-Person Tracking-by-Detection from a
  Single, Uncalibrated Camera}}.
\newblock \bibinfo{journal}{IEEE Transactions on Pattern Analysis and Machine
  Intelligence (PAMI)} \bibinfo{volume}{33}, \bibinfo{pages}{1820--1833}.
\bibitem[{Daamen and Hoogendoorn(2012)}]{Daamen2012a}
\bibinfo{author}{Daamen, W.}, \bibinfo{author}{Hoogendoorn, S.P.},
  \bibinfo{year}{2012}.
\newblock \bibinfo{title}{{Calibration of Pedestrian Simulation Model for
  Emergency Doors for Different Pedestrian Types}}, in:
  \bibinfo{booktitle}{Proceedings of the Transportation Research Board 91st
  Annual Meeting (TRB 2012)}, \bibinfo{address}{Washington D. C., USA}.
\bibitem[{Duda et~al.(2001)Duda, Hart and Stork}]{DudaandHart}
\bibinfo{author}{Duda, R.}, \bibinfo{author}{Hart, P.}, \bibinfo{author}{Stork,
  D.}, \bibinfo{year}{2001}.
\newblock \bibinfo{title}{{Pattern Classification}}.
\newblock \bibinfo{publisher}{Wiley}.
\bibitem[{Eiter and Mannila(1994)}]{Wien94computingdiscrete}
\bibinfo{author}{Eiter, T.}, \bibinfo{author}{Mannila, H.},
  \bibinfo{year}{1994}.
\newblock \bibinfo{title}{{Computing Discrete Fr\'{e}chet Distance}}.
\newblock \bibinfo{type}{Technical Report} \bibinfo{number}{CD-TR 94/64}.
  Vienna University of Technology.
\bibitem[{Forsyth and Ponce(2002)}]{Forsyth2002}
\bibinfo{author}{Forsyth, D.A.}, \bibinfo{author}{Ponce, J.},
  \bibinfo{year}{2002}.
\newblock \bibinfo{title}{{Computer Vision: A Modern Approach}}.
\newblock \bibinfo{publisher}{Prentice Hall}. \bibinfo{edition}{1} edition.
\bibitem[{Frati and Prattichizzo(2011)}]{Frati2011}
\bibinfo{author}{Frati, V.}, \bibinfo{author}{Prattichizzo, D.},
  \bibinfo{year}{2011}.
\newblock \bibinfo{title}{{Using Kinect for Hand Tracking and Rendering in
  Wearable Haptics}}, in: \bibinfo{booktitle}{Proceedings of the IEEE World
  Haptics Conference (WHC 2011)}, pp. \bibinfo{pages}{317 --321}.
\bibitem[{Girshick et~al.(2011)Girshick, Shotton, Kohli, Criminisi and
  Fitzgibbon}]{Girshick2011}
\bibinfo{author}{Girshick, R.}, \bibinfo{author}{Shotton, J.},
  \bibinfo{author}{Kohli, P.}, \bibinfo{author}{Criminisi, A.},
  \bibinfo{author}{Fitzgibbon, A.}, \bibinfo{year}{2011}.
\newblock \bibinfo{title}{{Efficient Regression of General-Activity Human Poses
  from Depth Images}}, in: \bibinfo{booktitle}{Proceedings of the IEEE
  International Conference on Computer Vision (ICCV 2011)},
  \bibinfo{publisher}{IEEE Computer Society}, \bibinfo{address}{Los Alamitos,
  CA, USA}. pp. \bibinfo{pages}{415 --422}.
\bibitem[{Helbing and Johansson(2009)}]{Helbing2010}
\bibinfo{author}{Helbing, D.}, \bibinfo{author}{Johansson, A.},
  \bibinfo{year}{2009}.
\newblock \bibinfo{title}{{Pedestrian, Crowd and Evacuation Dynamics}}.
\newblock \bibinfo{journal}{Encyclopedia of Complexity and Systems Science}
  \bibinfo{volume}{16}, \bibinfo{pages}{6476--6495}.
\bibitem[{Helbing and Moln\'ar(1995)}]{Helbing1995}
\bibinfo{author}{Helbing, D.}, \bibinfo{author}{Moln\'ar, P.},
  \bibinfo{year}{1995}.
\newblock \bibinfo{title}{{Social Force Model for Pedestrian Dynamics}}.
\newblock \bibinfo{journal}{Physical Review E} \bibinfo{volume}{51},
  \bibinfo{pages}{4282--4286}.
\bibitem[{Hoogendoorn and Daamen(2006)}]{Hoogendoorn2006}
\bibinfo{author}{Hoogendoorn, S.}, \bibinfo{author}{Daamen, W.},
  \bibinfo{year}{2006}.
\newblock \bibinfo{title}{{Microscopic Parameter Identification of Pedestrian
  Models and Implications for Pedestrian Flow Modelling}}.
\newblock \bibinfo{journal}{Transportation Research Record}
  \bibinfo{volume}{1982}, \bibinfo{pages}{57--64}.
\bibitem[{Hoogendoorn and Daamen(2003)}]{Hoogendoorn2003a}
\bibinfo{author}{Hoogendoorn, S.P.}, \bibinfo{author}{Daamen, W.},
  \bibinfo{year}{2003}.
\newblock \bibinfo{title}{{Extracting Microscopic Pedestrian Characteristics
  from Video Data}}, in: \bibinfo{booktitle}{Proceedings of the Transportation
  Research Board 82st Annual Meeting (TRB 2003)}, \bibinfo{address}{Washington
  D. C., USA}. pp. \bibinfo{pages}{1--15}.
\bibitem[{Hoogendoorn and Daamen(2005)}]{Hoogendoorn2005}
\bibinfo{author}{Hoogendoorn, S.P.}, \bibinfo{author}{Daamen, W.},
  \bibinfo{year}{2005}.
\newblock \bibinfo{title}{{Pedestrian Behavior at Bottlenecks}}.
\newblock \bibinfo{journal}{Transportation Science} \bibinfo{volume}{39},
  \bibinfo{pages}{147--159}.
\bibitem[{Izadi et~al.(2011)Izadi, Kim, Hilliges, Molyneaux, Newcombe, Kohli,
  Shotton, Hodges, Freeman, Davison and Fitzgibbon}]{Izadi2011}
\bibinfo{author}{Izadi, S.}, \bibinfo{author}{Kim, D.},
  \bibinfo{author}{Hilliges, O.}, \bibinfo{author}{Molyneaux, D.},
  \bibinfo{author}{Newcombe, R.}, \bibinfo{author}{Kohli, P.},
  \bibinfo{author}{Shotton, J.}, \bibinfo{author}{Hodges, S.},
  \bibinfo{author}{Freeman, D.}, \bibinfo{author}{Davison, A.},
  \bibinfo{author}{Fitzgibbon, A.}, \bibinfo{year}{2011}.
\newblock \bibinfo{title}{{KinectFusion: Real-Time 3D Reconstruction and
  Interaction Using a Moving Depth Camera}}, in:
  \bibinfo{booktitle}{Proceedings of the 24th Annual ACM Symposium on User
  Interface Software and Technology (UIST 2011)}, \bibinfo{publisher}{ACM},
  \bibinfo{address}{New York, NY, USA}. pp. \bibinfo{pages}{559--568}.
\bibitem[{Johansson and Helbing(2010)}]{Johansson2008}
\bibinfo{author}{Johansson, A.}, \bibinfo{author}{Helbing, D.},
  \bibinfo{year}{2010}.
\newblock \bibinfo{title}{{Analysis of Empirical Trajectory Data of
  Pedestrians}}, in: \bibinfo{editor}{Klingsch, W.W.F.},
  \bibinfo{editor}{Rogsch, C.}, \bibinfo{editor}{Schadschneider, A.},
  \bibinfo{editor}{Schreckenberg, M.} (Eds.), \bibinfo{booktitle}{Proceedings
  of the Conference on Pedestrian and Evacuation Dynamics (PED 2008)},
  \bibinfo{publisher}{Springer}, \bibinfo{address}{Berlin, Heidelberg}. pp.
  \bibinfo{pages}{203--214}.
\bibitem[{Kaucic et~al.(2005)Kaucic, Amitha~Perera, Brooksby, Kaufhold and
  Hoogs}]{Kaucic2005}
\bibinfo{author}{Kaucic, R.}, \bibinfo{author}{Amitha~Perera, A.},
  \bibinfo{author}{Brooksby, G.}, \bibinfo{author}{Kaufhold, J.},
  \bibinfo{author}{Hoogs, A.}, \bibinfo{year}{2005}.
\newblock \bibinfo{title}{{A Unified Framework for Tracking through Occlusions
  and across Sensor Gaps}}, in: \bibinfo{booktitle}{Proceedings of the IEEE
  Conference on Computer Vision and Pattern Recognition (CVPR 2005)}, pp.
  \bibinfo{pages}{990 -- 997 vol. 1}.
\bibitem[{Lagarias et~al.(1998)Lagarias, Reeds, Wright and
  Wright}]{Lagarias1998}
\bibinfo{author}{Lagarias, J.C.}, \bibinfo{author}{Reeds, J.A.},
  \bibinfo{author}{Wright, M.H.}, \bibinfo{author}{Wright, P.E.},
  \bibinfo{year}{1998}.
\newblock \bibinfo{title}{{Convergence Properties of the Nelder-Mead Simplex
  Method in Low Dimensions}}.
\newblock \bibinfo{journal}{SIAM Journal of Optimization} \bibinfo{volume}{9},
  \bibinfo{pages}{112--147}.
\bibitem[{Massey(1951)}]{Massey51}
\bibinfo{author}{Massey, F.J.}, \bibinfo{year}{1951}.
\newblock \bibinfo{title}{{The {K}olmogorov-{S}mirnov Test for Goodness of
  Fit}}.
\newblock \bibinfo{journal}{Journal of the American Statistical Association}
  \bibinfo{volume}{46}, \bibinfo{pages}{68--78}.
\bibitem[{{Microsoft Corp.}(2012a)}]{MicrosoftKinect}
\bibinfo{author}{{Microsoft Corp.}}, \bibinfo{year}{2012}a.
\newblock \bibinfo{title}{{Kinect for Xbox 360}}.
\newblock \bibinfo{address}{{Redmond, WA, USA}}.
\bibitem[{{Microsoft Corp.}(2012b)}]{KinectSDK}
\bibinfo{author}{{Microsoft Corp.}}, \bibinfo{year}{2012}b.
\newblock \bibinfo{title}{{Microsoft Kinect for Windows SDK}}.
\newblock
  \bibinfo{howpublished}{\url{http://www.microsoft.com/en-us/kinectforwindows/}}.
\newblock \bibinfo{note}{(accessed August 2012)}.
\bibitem[{Munkres(1957)}]{Munkres1957}
\bibinfo{author}{Munkres, J.}, \bibinfo{year}{1957}.
\newblock \bibinfo{title}{{Algorithms for the Assignment and Transportation
  Problems}}.
\newblock \bibinfo{journal}{Journal of the Society of Industrial and Applied
  Mathematics} \bibinfo{volume}{5}, \bibinfo{pages}{32--38}.
\bibitem[{Noonan et~al.(2011)Noonan, Cootes, Hallett and Hinz}]{Noonan2011}
\bibinfo{author}{Noonan, P.J.}, \bibinfo{author}{Cootes, T.F.},
  \bibinfo{author}{Hallett, W.A.}, \bibinfo{author}{Hinz, R.},
  \bibinfo{year}{2011}.
\newblock \bibinfo{title}{{The Design and Initial Calibration of an Optical
  Tracking System using the Microsoft Kinect}}, in:
  \bibinfo{booktitle}{Proceedings of the IEEE Nuclear Science Symposium and
  Medical Imaging Conference (NSS/MIC 2011)}, pp. \bibinfo{pages}{3614--3617}.
\bibitem[{Plaue et~al.(2011)Plaue, Chen, B\"{a}rwolff and Schwandt}]{Plaue2011}
\bibinfo{author}{Plaue, M.}, \bibinfo{author}{Chen, M.},
  \bibinfo{author}{B\"{a}rwolff, G.}, \bibinfo{author}{Schwandt, H.},
  \bibinfo{year}{2011}.
\newblock \bibinfo{title}{{Trajectory Extraction and Density Analysis of
  Intersecting Pedestrian Flows from Video Recordings}}, in:
  \bibinfo{booktitle}{Proceedings of the 2011 ISPRS Conference on
  Photogrammetric Image Analysis (PIA 2011)}, \bibinfo{publisher}{Springer},
  \bibinfo{address}{Berlin, Heidelberg}. pp. \bibinfo{pages}{285--296}.
\bibitem[{Rudloff et~al.(2011)Rudloff, Matyus, Seer and Bauer}]{Rudloff2011}
\bibinfo{author}{Rudloff, C.}, \bibinfo{author}{Matyus, T.},
  \bibinfo{author}{Seer, S.}, \bibinfo{author}{Bauer, D.},
  \bibinfo{year}{2011}.
\newblock \bibinfo{title}{{Can Walking Behavior be Predicted? An Analysis of
  the Calibration and Fit of Pedestrian Models}}.
\newblock \bibinfo{journal}{Transportation Research Record}
  \bibinfo{volume}{2264}, \bibinfo{pages}{101--109}.
\bibitem[{Shotton et~al.(2011)Shotton, Fitzgibbon, Cook, Sharp, Finocchio,
  Moore, Kipman and Blake}]{Shotton2011}
\bibinfo{author}{Shotton, J.}, \bibinfo{author}{Fitzgibbon, A.},
  \bibinfo{author}{Cook, M.}, \bibinfo{author}{Sharp, T.},
  \bibinfo{author}{Finocchio, M.}, \bibinfo{author}{Moore, R.},
  \bibinfo{author}{Kipman, A.}, \bibinfo{author}{Blake, A.},
  \bibinfo{year}{2011}.
\newblock \bibinfo{title}{{Real-Time Human Pose Recognition in Parts from a
  Single Depth Image}}, in: \bibinfo{booktitle}{Proceedings of the IEEE
  Conference on Computer Vision and Pattern Recognition (CVPR 2011)},
  \bibinfo{publisher}{IEEE Computer Society}, \bibinfo{address}{Los Alamitos,
  CA, USA}. pp. \bibinfo{pages}{1297--1304}.
\bibitem[{Stauffer(2003)}]{Stauffer2003}
\bibinfo{author}{Stauffer, C.}, \bibinfo{year}{2003}.
\newblock \bibinfo{title}{{Estimating Tracking Sources and Sinks}}, in:
  \bibinfo{booktitle}{Proceedings of the IEEE Computer Vision and Pattern
  Recognition Workshop (CVPRW 2003)}, p.~\bibinfo{pages}{35}.
\bibitem[{Stauffer and Grimson(2000)}]{Stauffer2000}
\bibinfo{author}{Stauffer, C.}, \bibinfo{author}{Grimson, W.E.L.},
  \bibinfo{year}{2000}.
\newblock \bibinfo{title}{{Learning Patterns of Activity Using Real-Time
  Tracking}}.
\newblock \bibinfo{journal}{IEEE Transactions on Pattern Analysis and Machine
  Intelligence (PAMI)} \bibinfo{volume}{22}, \bibinfo{pages}{747--757}.
\bibitem[{Weiss et~al.(2011)Weiss, Hirshberg and Black}]{Weiss2011}
\bibinfo{author}{Weiss, A.}, \bibinfo{author}{Hirshberg, D.},
  \bibinfo{author}{Black, M.}, \bibinfo{year}{2011}.
\newblock \bibinfo{title}{{Home 3D Body Scans from Noisy Image and Range
  Data}}, in: \bibinfo{booktitle}{Proceedings of the IEEE International
  Conference on Computer Vision (ICCV 2011)}, \bibinfo{address}{Barcelona}. pp.
  \bibinfo{pages}{1951--1958}.

\end{thebibliography}
